%% file: egpaper_final.tex
\documentclass[10pt,twocolumn,letterpaper]{article}

\usepackage{cvpr}
\usepackage{times}
\usepackage{epsfig}
\usepackage{graphicx}
\usepackage{amsmath}
\usepackage{amssymb}
\usepackage{color}
\usepackage{multicol, blindtext}
\usepackage{booktabs}
\usepackage{soul}
\usepackage{csquotes}
\usepackage[dvipsnames]{xcolor}
\definecolor{mypurple}{RGB}{160, 48, 160}
\definecolor{mygreen}{RGB}{84, 130, 53}
\definecolor{myorange}{RGB}{255, 210, 0}
\definecolor{myblue}{RGB}{0, 176, 240}
\definecolor{mybrown}{RGB}{190, 62, 19}
\newcommand{\RomanNumeralCaps}[1]
    {\MakeUppercase{\romannumeral #1}}

\usepackage[breaklinks=true,bookmarks=false]{hyperref}
\cvprfinalcopy 
\def\cvprPaperID{****} 

\begin{document}

\title{Dual Contrastive Learning for Unsupervised Image-to-Image Translation}


\author{Junlin Han$^{1,2}$ \quad
Mehrdad Shoeiby$^{1}$ \quad
Lars Petersson$^{1}$ \quad
Mohammad Ali Armin$^{1}$\\
$^{1}$DATA61-CSIRO, $^{2}$Australian National University\\
{\tt\small\{junlin.han, mehrdad.shoeiby, lars.petersson, ali.armin\}@data61.csiro.au}
}

\maketitle
\begin{abstract}
Unsupervised image-to-image translation tasks aim to find a mapping between a source domain X and a target domain Y from unpaired training data. Contrastive learning for Unpaired image-to-image Translation (CUT) yields state-of-the-art results in modeling unsupervised image-to-image translation by maximizing mutual information between input and output patches using only one encoder for both domains. In this paper, we propose a novel method based on contrastive learning and a dual learning setting (exploiting two encoders) to infer an efficient mapping between unpaired data. Additionally, while CUT suffers from mode collapse, a variant of our method efficiently addresses this issue. We further demonstrate the advantage of our approach through extensive ablation studies demonstrating superior performance comparing to recent approaches in multiple challenging image translation tasks. Lastly, we demonstrate that the gap between unsupervised methods and supervised methods can be efficiently closed.
\end{abstract}

\section{Introduction}
The image-to-image translation task aims to convert images from one domain to another domain, e.g., horse to zebra, low-resolution images to high-resolution images, image to label, photography to painting, and vice versa. Image-to-image translation has drawn considerable attention due to its wide range of applications including style-transfer~\cite{CycleGAN2017,huang2018multimodal,lee2018diverse,park2020cut,distance}, image in-painting~\cite{pathak2016context}, colourisation~\cite{zhang2016colorful},  super-resolution~\cite{kim2016accurate,yuan2018unsupervised}, dehazing~\cite{li2018single}, underwater image restoration~\cite{han2021cwr}, and denoising~\cite{chen2018image}.

In unsupervised image-to-image translation without paired data, the main problem is that the adversarial loss~\cite{goodfellow2014generative} is significantly under-constrained, that is, there exist multiple possible mappings between the two domains which make the training unstable and, hence, the translation unsuccessful. To restrict the mapping, the contemporary approaches CycleGAN~\cite{CycleGAN2017}, DiscoGAN~\cite{kim2017disco}, and DualGAN~\cite{yi2017dualgan} use a similar idea, the assumption of cycle-consistency~\cite{CycleGAN2017} which learns the reverse mapping from the target domain back to the source domain and measures whether the reconstruction image is identical to the input image. The cycle-consistency~\cite{CycleGAN2017} assumption ensures that the translated images have similar texture information to the target domain, failing to perform geometry changes. Also, the cycle-consistency~\cite{CycleGAN2017} assumption forces the relationship between the two domains to be a bijection~\cite{li2017alice}. This is usually not ideal. For example, in the horse to zebra image translation task, the reconstruction is constrained via a fidelity loss, compromising image diversity.

To address this constraint, recently contrastive learning between multiple views of the data has achieved state-of-the-art performance~\cite{he2020momentum,chen2020simple,henaff2019data,oord2018representation} in the field of self-supervised representation learning. This was followed by CUT~\cite{park2020cut} introducing contrastive learning for unpaired image-to-image translation with a patch-based, multi-layer PatchNCE loss to maximize the mutual information between corresponding patches of input and output images.

While CUT~\cite{park2020cut} demonstrated the efficiency of contrastive learning, we believe certain design choices are limiting its performance. For example, one embedding was used for two distinct domains which may not efficiently capture the domain gap. To further leverage contrastive learning and avoid the drawbacks of cycle-consistency~\cite{CycleGAN2017}, we propose our dual contrastive learning approach which is referred to as DCLGAN.
\input{ figs/DCLGAN}

DCLGAN aims to maximize mutual information by learning the correspondence between input and output image patches using separate embeddings. By employing different encoders and projection heads for different domains, we learn suitable embeddings to maximize agreement. The dual learning setting~\cite{yi2017dualgan} also helps to stabilize training. Besides, we revisit some design choices and find that removing RGB pixels representing small patches, in the PatchNCE loss, can be beneficial. We show that cycle-consistency~\cite{CycleGAN2017} is unnecessary and in fact counter-intuitive when there is no strict constraint on geometrical structure. Lastly, a variant of DCLGAN, referred to as SimDCL, significantly avoids mode collapse.

This paper presents a novel framework and its' variant that can break the limitations of CycleGAN~\cite{CycleGAN2017} (limited performance in geometry changes) and CUT~\cite{park2020cut} (suffering mode collapse and a few inferior results).
Through extensive experiments, we demonstrate the quantitative and qualitative superiority of our method compared to several state-of-the-art methods on various popular tasks. Additionally, we show that our method successfully closes the gap between unsupervised and supervised methods, as contrastive learning has done in the field of self-supervised learning. A comprehensive ablation study demonstrates the effectiveness of DCLGAN. Our code is available at \textcolor{red}{\href{https://github.com/JunlinHan/DCLGAN}{GitHub}}.

\section{Related Work}
\paragraph{Image-to-image Translation.} GANs~\cite{goodfellow2014generative} have been applied to a multitude of image applications, especially in image-to-image translation. The key to the success of GANs is the idea of adversarial loss~\cite{goodfellow2014generative}, which forces the generated image to be indistinguishable in principle from the real image. Generally, image-to-image translation can be categorized into two groups: a paired setting (supervised)~\cite{wang2018pix2pixHD,isola2017image,park2019SPADE} and an unpaired setting (unsupervised)~\cite{CycleGAN2017,kim2017disco,yi2017dualgan}. Paired setting means the training set is paired, every image from domain $X$ has a corresponding image from domain $Y$.

\textbf{Supervised methods.}
In this line, Pix2Pix~\cite{isola2017image} first achieved task-agnostic image translation supporting multiple image-to-image translation tasks using only a general method. It has then been extended to Pix2PixHD~\cite{wang2018pix2pixHD} enabling synthesizing high-resolution photo-realistic images. SPADE~\cite{park2019SPADE} introduces the spatially-adaptive normalization layer to further improve the quality of generated images. These supervised approaches require paired data for training, which imposes a limitation on their usage.

\textbf{Unsupervised methods.}
In unsupervised settings, the current  methods~\cite{CycleGAN2017,huang2018multimodal,park2020cut,lee2018diverse,choi2020stargan,yi2017dualgan,distance,kim2017disco,geometry2019,liu2017unsupervised,lee2020drit++,yi2017dualgan,chen2020reusing,zhao2020aclgan} are mainly developed based on two assumptions: a shared latent space~\cite{liu2017unsupervised} and a cycle-consistency assumption~\cite{CycleGAN2017}. UNIT~\cite{liu2017unsupervised} proposes a shared latent space assumption which assumes a pair of corresponding images in different domains can be mapped to the same latent representation in a shared-latent space.
Recent works~\cite{huang2018multimodal,lee2018diverse,choi2020stargan,lee2020drit++,choi2018stargan} further enable multi-modal and multi-domain synthesis to bring diversity in the translated outputs. MUNIT~\cite{huang2018multimodal} disentangles domain-specific features by splitting the latent space into style code and content code. DRIT~\cite{lee2018diverse,lee2020drit++} embeds images onto two spaces including a domain-specific attribute space and a content space capturing shared information. StarGAN~\cite{choi2018stargan,choi2020stargan} employs a unified model architecture to translate images across multiple domains.

\textbf{Break the cycle.}
CycleGAN~\cite{CycleGAN2017} learns two mappings simultaneously via translating an image to the target domain and back preserving the fidelity of the input and the reconstructed image. This leads it to be too restrictive. Recently, a few methods~\cite{park2020cut,nizan2020council,distance,geometry2019} have tried to break the cycle to alleviate the problem of cycle-consistency~\cite{CycleGAN2017}. CouncilGAN~\cite{nizan2020council} uses more than two generators and discriminators along with the council loss. DistanceGAN~\cite{distance} and GCGAN~\cite{geometry2019} enable one-way translation. They employ different constraints from different aspects. We take the advantages of both CycleGAN~\cite{CycleGAN2017} and CUT~\cite{park2020cut}, employing the idea of mutual information maximization to enable two-sided unsupervised image-to-image translation based on the architectures of CycleGAN~\cite{CycleGAN2017}.

\textbf{Contrastive learning.}
In the field of unsupervised representation learning~\cite{chen2020simple,he2020momentum,henaff2019data,oord2018representation}, contrastive learning aims to learn an embedding where associated signals are pulled together while other samples in the dataset are pushed away. Signals may vary depending on specific tasks. In general, the objective is to discriminate its transformed version against other samples. To get the transformed version, data augmentation shows the most successful results ~\cite{chen2020simple,he2020momentum}. However, natural transformations can draw comparable results~\cite{henaff2019data,Han20} if ideal natural sources such as those arising audio and optical flow in videos are available. For image-to-image translations, patches are ideal natural sources for instance discrimination since they are easy to track and use~\cite{henaff2019data,oord2018representation,park2020cut}.
CUT~\cite{park2020cut} first applies noise contrastive estimation to image-to-image translation tasks by learning the correspondence between input image patches and the corresponding generated image patches, achieving a performance superior to those based on cycle-consistency~\cite{CycleGAN2017}. We further rethink several design choices of leveraging contrastive learning, making it more beneficial to employ contrastive learning for unsupervised image-to-image translation. We extend one-sided mapping to two-sided, performing better in learning embeddings and thus achieving new state-of-the-art results. We additionally address the mode collapse problem that previous methods based on mutual information maximization can not handle.
\section{Method}
Given two domains $\mathcal{X} \subset \mathbb{R}^{H \times W \times C}$ and $\mathcal{Y} \subset \mathbb{R}^{H \times W \times 3}$ and a dataset of unpaired instances $X$ containing some images $x$ and $Y$ containing some images $y$. We denote them $X= \left\{x \in \mathcal{X} \right\}$ and $Y= \left\{y \in \mathcal{Y} \right\}$. We aim to learn two mappings $G : X\rightarrow Y$ and $F : Y\rightarrow X$.

DCLGAN has two generators $G, F$ as well as two discriminators $D_{X}$, $D_{Y}$. $G$ enables the mapping from domain $X$ to domain $Y$ and $F$ enables the reverse mapping. $D_{X}$ and $D_{Y}$ ensure that the translated images belong to the correct image domain. The first half of the generators are defined as encoder while the second half are decoders and presented as $G_{enc}$ and $F_{enc}$ followed by $G_{dec}$ and $F_{dec}$ respectively.

For each mapping, we extract features of images from four layers of the encoder and send them to a two-layer MLP projection head ($H_{X}$ and $H_{Y}$). Such a projection head learns to project the extracted features from the encoder to a stack of features. Note that we use $G_{enc}$ and $H_{X}$ as the embedding for domain $X$ and use $F_{enc}$ and $H_{Y}$ as the embedding for domain $Y$. If two domains share common semantic information such as horse and zebra, using one encoder can provide reasonable results. However, this may fail to capture the variability in two distinctive domains with a large gap. Additionally, we introduce four light networks to capture the common information within one domain and form a similarity loss.

Figure~\ref{fig:DCLGAN} shows the overall architecture of DCLGAN and SimDCL. DCLGAN combines three losses including adversarial loss~\cite{goodfellow2014generative}, PatchNCE loss, and identity loss~\cite{CycleGAN2017} whereas SimDCL has one additional similarity loss to address mode collapse. The details of our objective are described below.

\subsection{Adversarial loss}
An adversarial loss~\cite{goodfellow2014generative} is employed to encourage the generator to generate visually similar images to images from the target domain, for the mapping $G : X\rightarrow Y$ with discriminator $D_{Y}$ , the GAN loss is calculated by:
\begin{equation}
\begin{aligned}
\mathcal{L}_{\mathrm{GAN}}\left(G, D_{Y}, X, Y\right) &=\mathbb{E}_{y\sim{Y}}\left[\log D_{Y}(y)\right] \\
&+\mathbb{E}_{x\sim{X}}\left[\log \left(1-D_{Y}(G(x))\right]\right.,
\end{aligned}
\end{equation}
where $G$ tries to generate images $G(x)$ that look similar to
images from domain $Y$ , while $D_{Y}$ aims to distinguish between translated samples $G(x)$ and real samples $y$. A similar adversarial loss for the mapping
 $F : Y\rightarrow X$ and its discriminator $D_{X}$ is introduced as:
\begin{equation}
\begin{aligned}
\mathcal{L}_{\mathrm{GAN}}\left(F, D_{X}, X, Y\right) &=\mathbb{E}_{x\sim{X}}\left[\log D_{X}(x)\right] \\
&+\mathbb{E}_{y\sim{Y}}\left[\log \left(1-D_{X}(G(y))\right]\right..
\end{aligned}
\end{equation}

\subsection{Patch-based multi-layer contrastive learning}
\paragraph{Mutual information maximization.}
Our goal is to maximize the mutual information between corresponding patches of the input and the output. For instance, for a patch showing the eye of a generated dog (top-right of Figure~\ref{fig:DCLGAN}), we should be able to more strongly associate it with the eye of the input real cat other than the rest of the patches of the cat.

Following the setting of CUT~\cite{park2020cut}, we employ a noisy contrastive estimation framework~\cite{oord2018representation} to maximize the mutual information between inputs and outputs. The idea behind contrastive learning is to correlate two signals, i.e., the ``query'' and its' ``positive'' example, in contrast to other examples in the dataset (referred to as ``negatives'').

We map query, positive, and $N$ negatives to $K$-dimensional vectors and denote them $v, v^{+} \in R^{K}$ and $ v^{-} \in R^{N \times K}$ respectively. Note that $v_{n}^{-} \in R^{K}$ denotes the n-th negative. We normalize vectors with L2-normalization then set up an $(N+1)$-way classification problem and compute the probability that a ``positive'' is selected over ``negatives''. Mathematically, this can be expressed as a cross-entropy loss~\cite{gutmann2010noise} which is computed by:
\begin{equation}
\begin{aligned}
    &\ell\left(\boldsymbol{v}, \boldsymbol{v}^{+}, \boldsymbol{v}^{-}\right)= -\log( \\
    &\frac{\exp \left(\boldsymbol sim({v},\boldsymbol{v}^{+}) / \tau\right)}{\exp \left(\boldsymbol sim({v},\boldsymbol{v}^{+}) / \tau\right)+\sum_{n=1}^{N} \exp \left(\boldsymbol sim({v},\boldsymbol{v}_{n}^{-}) / \tau\right)}),
\end{aligned}
\label{eq:nce}
\end{equation}
where $\operatorname{sim}(\boldsymbol{u}, \boldsymbol{v})=\boldsymbol{u}^{\top} \boldsymbol{v} /\|\boldsymbol{u}\|\|\boldsymbol{v}\|$ denotes the cosine similarity between $\boldsymbol{u}$ and $\boldsymbol{v}$. $\tau$ denotes a temperature parameter to scale the distance between the query and other examples, we use 0.07 as default.

\textbf{PatchNCE loss.}
We use $G_{enc}$ and $H_{X}$ to extract features from domain $X$ and use $F_{enc}$ and $H_{Y}$ to extract features from domain $Y$. We do not share weights in order to learn better embeddings and capture variability in two distinct domains.
We select $L$ layers from $G_{enc}(X)$ and send it to $H_{X}$, embedding one image to a stack of features $\left\{\boldsymbol{z}_{l}\right\}_{L}=\left\{H_{X}^{l}\left(G_{\mathrm{enc}}^{l}(\boldsymbol{x})\right)\right\}_{L}$, where $G_{\mathrm{enc}}^{l}$ represents the output of $l$-th selected layers.

Now we consider the patches. After having a stack of features, each feature actually represents one patch from the image. We take advantage of that and denote the spatial locations in each selected layer as $s \in \{1,...,S_{l}\}$, where $S_{l}$ is the number of spatial locations in each layer.
We select a query each time, refer the corresponding feature (``positive'') as $\boldsymbol{z}_{l}^{s} \in \mathbb{R}^{C_{l}}$ and all other features ``negatives'') as $\boldsymbol{z}_{l}^{S \backslash s} \in \mathbb{R}^{\left(S_{l}-1\right) \times C_{l}}$, where $C_{l}$ is the number of channels in each layer.
For the generated fake image $G(x)$ belonging to domain $Y$, we exploit the advantages of dual learning and use a different embedding of domain $Y$. Similarly, we get another stack of features $\left\{\hat{\boldsymbol{z}}_{l}\right\}_{L}=\left\{H_{Y}^{l}\left(F_{\mathrm{enc}}^{l}(G(\boldsymbol{x}))\right)\right\}_{L}$.

We aim to match the corresponding patches of input and output images. The patch-based, multi-layer PatchNCE loss~\cite{park2020cut} for mapping $G : X\rightarrow Y$ can be expressed as:
\begin{equation}
\begin{aligned}
&\mathcal{L}_{\mathrm{PatchNCE_{X}}}(G,H_{X},H_{Y},X)=\\
&\mathbb{E}_{\boldsymbol{x} \sim X} \sum_{l=1}^{L} \sum_{s=1}^{S_{l}} \ell\left(\hat{z}_{l}^{s}, \boldsymbol{z}_{l}^{s}, \boldsymbol{z}_{l}^{S \backslash s}\right).
\end{aligned}
\label{eq:patchnce}
\end{equation}
Consider the reverse mapping $F : Y\rightarrow X$, we introduce a similar loss as well,
\begin{equation}
\begin{aligned}
&\mathcal{L}_{\mathrm{PatchNCE_{Y}}}(F,H_{X},H_{Y},Y)=\\
&\mathbb{E}_{\boldsymbol{y} \sim Y} \sum_{l=1}^{L} \sum_{s=1}^{S_{l}} \ell\left(\hat{z}_{l}^{s}, \boldsymbol{z}_{l}^{s}, \boldsymbol{z}_{l}^{S \backslash s}\right),
\end{aligned}
\label{eq:patchnce2}
\end{equation}
where $\left\{\boldsymbol{z}_{l}\right\}_{L}=\left\{H_{Y}^{l}\left(F_{\mathrm{enc}}^{l}(\boldsymbol{y})\right)\right\}_{L}$ and  $\left\{\hat{\boldsymbol{z}}_{l}\right\}_{L}=\left\{H_{X}^{l}\left(G_{\mathrm{enc}}^{l}(F(\boldsymbol{y}))\right)\right\}_{L}$ are different from $G : X\rightarrow Y$ .
\subsection{Similarity loss}
Intuitively, images from the same domain should have some similarities. Their semantics are different but they share a common style. In the dual learning setting, we have one real and one fake image belonging to the same domain in each iteration. After getting four stacks of features, we use four light networks ($H_{xr}, H_{xf}, H_{yr}, H_{yf}$) to project them to 64-dim vectors, where x, y, r, f refers to images within domain X, images within domain Y, real, fake correspondingly.
These 64-dim vectors belonging to the same domain can be measured by a similarity loss, such a loss can be formalised as:
\begin{equation}
\begin{aligned}
&\mathcal{L}_{\text {sim}}(G,F,H_{X},H_{Y},H_{xr},H_{xf},H_{yr},H_{yf})\\
&=\left[\|H_{xr}(H_{X}(G_{enc}(x)))-H_{xf}(H_{X}(G_{enc}(F(y))))\|_{1}^{sum}\right]\\
&+\left[\|H_{yr}(H_{Y}(F_{enc}(y)))-H_{yf}(H_{Y}(F_{enc}(G(x))))\|_{1}^{sum}\right],
\end{aligned}
\end{equation}
where $sum$ means we sum them up together. Implementing a similarity loss on the deep features forces the generated images to be realistic, as opposed to mode collapsed outputs, by encouraging the deep features of the generated and real images to be similar.
\subsection{Identity loss}
In order to prevent generators from unnecessary changes, we add an identity loss~\cite{CycleGAN2017}. Unlike CUT~\cite{park2020cut}, We do not employ PatchNCE loss as identity loss due to training speed.
\begin{equation}
\begin{aligned}
    \mathcal{L}_{\text {identity}}(G, F)&=\mathbb{E}_{x \sim {X} }\left[\|F(x)-x\|_{1}\right]\\
    &+\mathbb{E}_{y \sim {Y} }\left[\|G(y)-y\|_{1}\right].
\end{aligned}
\end{equation}
Such an identity loss can encourage the mappings to preserve color composition between the input and output.
\subsection{General objective}
\paragraph{DCLGAN.}
The generated image should be realistic and patches in the input and output images should share same correspondence. We employ identity loss~\cite{CycleGAN2017} in the default setting. The full objective is estimated by:
\begin{equation}
\begin{aligned}
&\mathcal{L}(G,F,D_{X},D_{Y},H_{X},H_{Y})\\
&=\lambda_{GAN}(\mathcal{L}_{GAN}(G,D_{Y},X,Y) +  \mathcal{L}_{GAN}(F,D_{X},X,Y)) \\
&+\lambda_{NCE}\mathcal{L}_{\mathrm{PatchNCE_{X}}}(G, H_{X}, H_{Y}, X)\\
&+\lambda_{NCE}\mathcal{L}_{\mathrm{PatchNCE_{Y}}}(F, H_{X}, H_{Y}, Y)\\
&+\lambda_{idt}\mathcal{L}_{\text {identity}}(G, F).
\end{aligned}
\end{equation}
We set $\lambda_{GAN}$ = 1, $\lambda_{NCE}$ = 2 and $\lambda_{idt}$ = 1. DCLGAN achieves superior performance to existing methods.

\textbf{SimDCL.}
We introduce SimDCL since methods based on mutual information maximization suffer from mode collapse in some specific tasks. We add similarity loss to the full objective of DCLGAN, and name it SimDCL, where sim is short for similarity and DCL stands for dual contrastive learning. The full objective of this variant is:
\begin{equation}
\begin{aligned}
&\mathcal{L}(G,F,D_{X},D_{Y},H_{X},H_{Y})\\
&=\lambda_{GAN}(\mathcal{L}_{GAN}(G,D_{Y},X,Y) + \mathcal{L}_{GAN}(F,D_{X},X,Y)) \\
&+\lambda_{NCE}\mathcal{L}_{\mathrm{PatchNCE_{X}}}(G, H_{X}, H_{Y}, X)\\
&+\lambda_{NCE}\mathcal{L}_{\mathrm{PatchNCE_{Y}}}(F, H_{X}, H_{Y}, Y)\\
&+\lambda_{sim}\mathcal{L}_{\text {sim}}(G,F,H_{X},H_{Y},H_{1},H_{2},H_{3},H_{4})\\
&+\lambda_{idt}\mathcal{L}_{\text {identity}}(G, F).
\end{aligned}
\end{equation}
We set $\lambda_{GAN}$ = 1, $\lambda_{NCE}$ = 2, $\lambda_{SIM}$ = 10 and $\lambda_{idt}$ = 1. This variant runs slower than DCLGAN, we recommend using it for Photo $\rightarrow$ Label, semantic segmentation and similar tasks to avoid mode collapse. SimDCL achieves equal or slightly worse performance compared to DCLGAN.
\input{ figs/comparisonfull}
\section{Experiments}
The training details, datasets, and our evaluation protocol along with all baselines are described as follows.
\subsection{Training details}
We mostly follow the setting of CUT~\cite{park2020cut} to train our proposed model. We use Hinge GAN loss~\cite{lim2017geometric} instead of LSGAN loss~\cite{mao2017least}. More specifically, we use the Adam optimiser~\cite{kingma2014adam} with $\beta_{1}$ = 0.5 and $\beta_{2}$ = 0.999. DCLGAN is trained for 400 epochs with a learning rate of 0.0001 while SimDCL is trained for 200 epochs with a learning rate of 0.0002 unless specified. The learning rate starts to decay linearly after half of the total epochs. We use a ResNet-based~\cite{he2016deep} generator with PatchGAN~\cite{isola2017image} as discriminator. We use a batch size of 1 and instance normalization~\cite{ulyanov2016instance}. All training images are loaded in $286 \times 286$ then cropped to $256 \times 256$ patches. More details on the training and the architecture are provided in the supplementary material.

\subsection{Datasets}
We evaluated our proposed method and baselines on six different datasets with nine tasks.

\textbf{Horse $\leftrightarrow$ Zebra} was introduced in CycleGAN, it contains 1067 horse images, 1344 zebra images as the training set and 260 test images all collected from ImageNet~\cite{deng2009imagenet}.

\textbf{Cat $\leftrightarrow$ Dog} contains 5000 training images and 500 test images for each domain. It was introduced in StarGAN2~\cite{choi2020stargan}. DCLGAN is trained for 200 epochs only for this dataset.

\textbf{CityScapes}~\cite{cordts2016cityscapes} contains 2975 training and 500 validation images for each domain. One domain is city scenes from German cities and the other is semantic segmentation labels. We focus on Label $\rightarrow$ City. We also leverage labels to measure how well methods discover correspondences.

\textbf{Van Gogh $\rightarrow$ Photo} contains 400 Van Gogh paintings and 6287 photographs from Flickr. It was collected in CycleGAN~\cite{CycleGAN2017}. DCLGAN is trained for 200 epochs only for this task. We reuse the training set of Van Gogh paintings as the test set.

\textbf{Label $\leftrightarrow$ Facade} is similar to CityScapes, it contains 400 paired training images and 106 paired test images from the CMP Facade Database~\cite{tylevcek2013spatial}.

\textbf{Orange $\rightarrow$ Apple} is also from ImageNet~\cite{deng2009imagenet}. It contains 1019 orange images and 995 apple images in the training set. For testing, we use 248 orange images.

\subsection{Evaluation}
\textbf{Metrics} We mainly use Fréchet Inception Distance (FID)~\cite{heusel2017gans} to measure the quality of generated images. FID~\cite{heusel2017gans} shows high correspondence with human perception, it is based on the Inception Score (IS)~\cite{salimans2016improved}. Lower FID~\cite{heusel2017gans} means lower Fréchet distance between real and generated images. That is to say, lower FID~\cite{heusel2017gans} means generated images are more realistic. For cityscapes, following Pix2Pix~\cite{isola2017image}, we use the pre-trained semantic segmentation network FCN-8~\cite{long2015fully} and compute three metrics. They are mean class Intersection over Union (IoU), pixel-wise accuracy (pixAcc), and average class accuracy (classAcc).

\textbf{Baselines.}
We perform qualitative and quantitative comparison between our proposed method and recent state-of-the-art unsupervised methods including CUT~\cite{park2020cut}, FastCUT~\cite{park2020cut}, CycleGAN~\cite{CycleGAN2017}, MUNIT~\cite{huang2018multimodal}, DRIT~\cite{lee2018diverse}, DistanceGAN~\cite{distance}, SelfDistance~\cite{distance} and GCGAN~\cite{geometry2019}. MUNIT~\cite{huang2018multimodal} and DRIT~\cite{lee2018diverse} are able to generate diverse results for only one input image and the others only produce one result. Among them, CUT, FastCUT~\cite{park2020cut}, DistanceGAN, SelfDistance~\cite{distance} and GcGAN~\cite{geometry2019} are one-sided methods. The rest are two-sided methods.

\section{Results}
Here, we compare our algorithms (DCLGAN and SimDCL) with all baselines on different datasets. Further, we compare DCLGAN to supervised methods on the CityScapes dataset using the FCN~\cite{long2015fully} score, showing that the performance of our method is on par with supervised methods. Lastly, we show that SimDCL avoids mode collapse.

\subsection{Comparison of different methods}
\input{ tables/comparision}

Table~\ref{tab:performance_comparison} shows a comparison of the quantitative results of DCLGAN and SimDCL with several baselines on three challenging tasks, including CityScapes, Cat $\rightarrow$ Dog, and Horse $\rightarrow$ Zebra. We only use the FID~\cite{heusel2017gans} score as our quantitative metric. It is evident that our algorithms perform stronger than all the baseline. Figure~\ref{fig:resultall} presents the corresponding \underline{randomly} selected qualitative results. DCLGAN performs both geometry changes and texture changes with negligible artifacts, this is especially successful in Cat $\rightarrow$ Dog while other methods can not generate realistic images. It is worth mentioning that models generating multiple outputs perform the worst.

\input{ figs/resultmore_double}
\input{ tables/comparisionmore.txt}

We select the top four methods from Table~\ref{tab:performance_comparison} and set a second comparison among them by testing them in 5 more tasks: Zebra $\rightarrow$ Horse, Van Gogh $\rightarrow$ Photo, Dog $\rightarrow$ Cat, Label $\rightarrow$ Facade and Orange $\rightarrow$ Dog. We show quantitative results in Table~\ref{tab:performance_more} and \underline{randomly} picked qualitative results in Figure~\ref{fig:second}. The results suggest that DCLGAN keeps superior performance comparing to other methods among various tasks. Methods under the cycle-consistency assumption~\cite{CycleGAN2017} usually fail to perform geometric changes while methods based on mutual information maximization successfully enable both geometric changes and texture changes. This is explicitly shown in Dog $\rightarrow$ Cat tasks.

\subsection{Comparison to supervised methods}
Here we compare our DCLGAN method with three popular supervised methods, Pix2Pix~\cite{isola2017image}, photo-realistic image synthesis system CRN~\cite{chen2017photographic} and discriminative region proposal adversarial network DRPAN~\cite{wang2018discriminative} on the CityScapes dataset. We follow the setting in Pix2Pix~\cite{isola2017image} and use a pre-trained semantic segmentation network FCN-8~\cite{long2015fully} to compute the FCN score. Quantitative results are shown in Table~\ref{tab:3compare}. DCLGAN performs best in pixACC and significantly closes the gap between unsupervised methods and supervised methods. On average, our method performs on par with supervised methods.
\input{ tables/3compare}
\input{ figs/facade}
\subsection{Addressing mode collapse via similarity loss.}
Our final comparison is a stress test on mode collapse. Mode collapse in generation tasks means the outputs lack diversity, and usually, the outputs are not realistic. We find that methods based on mutual information maximization (CUT and DCLGAN) can not prevent mode collapse in Photo $\rightarrow$ Label and similar tasks. To address this issue, we design SimDCL. We test the best four methods on the Facade $\rightarrow$ Label task and show the \underline{randomly} picked visual results in Figure~\ref{fig:facade}. No matter what the input is, the outputs of both CUT and DCLGAN are almost identical while SimDCL generates reasonable outputs for different inputs.
SimDCL is more robust to the mode collapse issue compared with other methods based on mutual information maximization.

\section{Ablation study}
\label{ablation}
DCLGAN shows superior performance compared to all baselines. We explore what is making contrastive learning effective. We analyze DCLGAN by studying each of our contributions in isolation via conducting several experiments, summarized in Table \ref{tab:ablation}. We use three tasks including Horse $\rightarrow$ Zebra, Zebra $\rightarrow$ Horse, and CityScapes in our ablation study.

We show the results of: (\RomanNumeralCaps{1}) Adding the first RGB pixels back. (\RomanNumeralCaps{2}) Drawing external negatives. (\RomanNumeralCaps{3}) Using the same encoder and MLP for one mapping instead of two. (\RomanNumeralCaps{4}) Adding cycle-consistency loss. (\RomanNumeralCaps{5}) Removing the dual setting.
\input{ tables/ablation}

\textbf{(\RomanNumeralCaps{1})} CUT~\cite{park2020cut} uses features from five layers in total including the first RGB pixels in PatchNCE loss ($l=5$ in Equations~\ref{eq:patchnce} and~\ref{eq:patchnce2}). Layers and spatial locations within the feature stack represent patches of the input image. Deeper layers correspond to bigger patches. However, RGB pixels represents the smallest possible patch size ($1\times 1$), providing misleading information. We find that not including the RGB layer encourages convergence. In fact, if we adopt the strategy in CUT~\cite{park2020cut} ($l=5$), the results deteriorate in all three tasks as demonstrated in Table \ref{tab:ablation}.

\textbf{(\RomanNumeralCaps{2}) Effect of drawing external negatives.} CUT~\cite{park2020cut} states that internal negatives (patches from an input image only) are more effective than external negatives (patches from other images). CUT~\cite{park2020cut} adds negatives using a momentum encoder~\cite{he2020momentum}.
We explore this in a different approach, by taking the advantage of the dual setting. DCLGAN produces four different stacks of features at each iteration. Concatenating two stacks of features belonging to the same domain provides more negatives (255 internal and 256 external) for one query while the default DCLGAN uses 255 internal negatives. We obverse better quantitative results in Horse $\rightarrow$ Zebra and very close results in CityScapes for this variant. Although the gap of FID score between the default DCLGAN and this variant is small, the visual quality is not as good as that of the default DCLGAN, that is, objects in the generated image tend to be merged together.

\textbf{(\RomanNumeralCaps{3}) Effect of using separate embeddings for each domain.}
While CUT~\cite{park2020cut} uses the same embedding for both domains, we use two separate embeddings, one for each domain. Adopting the CUT~\cite{park2020cut} strategy in our network we find that the results will deteriorate, as demonstrated in Table \ref{tab:ablation}. One embedding fails to capture the variability in two distinct domains, for instance, Photo $\rightarrow$ Label.

\textbf{(\RomanNumeralCaps{4}) Effect of Cycle-consistency loss.} To test if the cycle-consistency loss can improve the results, we add cycle-consistency~\cite{CycleGAN2017} loss to our objective. We did not observe any improvements (Table \ref{tab:ablation}). Although cycle-consistency and mutual information maximization share some commonalities, DCLGAN is much less restrictive. DCLGAN focuses on both texture and geometry changes while CycleGAN~\cite{CycleGAN2017} mostly focuses on texture only.
We tested this variant in two tasks requiring geometry changes, Cat $\rightarrow$ Dog and Dog $\rightarrow$ Cat, the FID scores are 71.1 and 35.5 respectively, all worse than the original DCLGAN. We conclude that when strict limitations on geometry are not crucial, cycle-consistency~\cite{CycleGAN2017} loss is better to be avoided.

\textbf{(\RomanNumeralCaps{5}) Dual settings stabilize the training.}
We remove the dual setting to demonstrate its' effect. We keep other settings the same as DCLGAN. The results are worse than DCLGAN, which shows the dual setting can learn better embeddings for different domains and stabilize the training.

\section{Conclusion}
We show that a dual setting can better leverage contrastive learning in unsupervised unpaired image-to-image translation. We also revise some significant designs to render contrastive learning more effective. In addition, a variant of DCLGAN, SimDCL mitigates mode collapse. Finally, we show that our method can hugely close the gap between unsupervised and supervised methods in challenging datasets such as CityScape, just as contrastive learning in the field of self-supervised representation learning.

{\small
\bibliographystyle{ieee_fullname}
\bibliography{egbib}
}

\clearpage
\appendix
\section{Appendix}
\subsection{Implementation Details}
\subsubsection{Architecture of Generator and layers used for PatchNCE loss}
Our generator architecture is based on CycleGAN~\cite{CycleGAN2017} and CUT~\cite{park2020cut}. We only use ResNet-based~\cite{he2016deep} generator with 9 residual blocks for training. It contains 2 downsampling blocks, 9 residual blocks, and 2 upsampling blocks. Each downsampling and upsampling block follows two-stride convolution/deconvolution, normalization, ReLU. Each residual block contains convolution, normalization, ReLU, convolution, normalization, and residual connection.

We define the first half of generators $G$ and $F$ as encoder which is represented as $G_{enc}$ and $F_{enc}$. The patch-based multi-layer PatchNCE loss is computed using features from four layers of the encoder (the first and second downsampling convolution, and the first and the fifth residual block). The patch sizes extracted from these four layers are 9×9, 15×15, 35×35, and 99×99 resolution respectively. Following CUT~\cite{park2020cut}, for each layer’s features, we sample 256 random locations and apply the 2-layer MLP (projection head $H_{X}, H_{Y}$) to infer 256-dim final features.

\subsubsection{Architecture of Discriminator}
We use the same PatchGAN discriminator architecture as CycleGAN~\cite{CycleGAN2017} and Pix2Pix~\cite{isola2017image} which uses local patches of sizes 70x70 and assigns every patch a result. This is equivalent to manually crop one image into 70x70 overlapping patches, run a regular discriminator over each patch, and average the results. For instance, the discriminator takes an image from either domain $X$ or domain $Y$, passes it through five downsampling Convolutional-Normalization-LeakeyReLU layers, and outputs a result matrix of 30x30. Each element corresponds to the classification result of one patch. Following CycleGAN~\cite{CycleGAN2017} and Pix2Pix~\cite{isola2017image}, in order to improve the stability of adversarial training, we use a buffer to store 50 previously generated images.

\subsubsection{Architecture of four light networks}
For SimDCL, we use four light networks ($H_{xr}, H_{xf}, H_{yr}, H_{yf}$). These networks project the 256-dim features to 64-dim vectors. Each network contains one convolutional layer followed by ReLU, average pooling, linear transformation (64-dim to 64-dim), ReLU, and linear transformation (64-dim to 64-dim).

\subsubsection{Additional training details}
We presented most training details in the main paper, here, we depict some additional training details.
For SimDCL, we use the Adam optimiser~\cite{kingma2014adam} with $\beta_{1}$ = 0.5 and $\beta_{2}$ = 0.999. We update the weights of $H_{X},H_{Y},H_{xr},H_{xf},H_{yr},H_{yf}$ together with learning rate 0.0002.

For both DCLGAN and SimDCL, we initialize weights using xavier initialization~\cite{glorot2010understanding}. We load all images in 286x286 resolution and randomly crop them into 256x256 patches during training and we load test images in 256x256 resolution. All images from the test set are used for evaluation. For all tasks, we train our method and other baselines with a Tesla P100-PCIE-16GB GPU. The GPU driver version is 440.64.00 and the CUDA version is 10.2.

\subsubsection{Additional evaluation details}
We list the evaluation details of Fréchet Inception Distance (FID)~\cite{heusel2017gans} and Fully convolutional Network (FCN)~\cite{long2015fully} score. For FID~\cite{heusel2017gans} score, we use the official PyTorch implementation with the default setting to match the evaluation protocol of CUT~\cite{park2020cut}. The link is \href{https://github.com/mseitzer/pytorch-fid}{https://github.com/mseitzer/pytorch-fid}.

For FCN~\cite{long2015fully} score, we use the official PyTorch implementation of CycleGAN~\cite{CycleGAN2017} and Pix2Pix~\cite{isola2017image}. The link is \href{https://github.com/junyanz/pytorch-CycleGAN-and-pix2pix}{https://github.com/junyanz/pytorch-CycleGAN-and-pix2pix}. The FCN~\cite{long2015fully} score is a well-known semantic segmentation metric on the CityScapes dataset. It measures how the algorithm finds correspondences between labels and images. The FCN~\cite{long2015fully} score is computed using a pre-trained FCN-8~\cite{long2015fully} network that predicts a label map for a photo. We input the generated photos to the pre-trained network and measure the predicted labels with ground truth using three semantic segmentation metrics including mean class Intersection over Union (IoU), pixel-wise accuracy (pixAcc), and average class accuracy (classAcc).

\subsection{Qualitative results of ablations}
For different ablations including (\RomanNumeralCaps{1}) Adding the first RGB pixels back, (\RomanNumeralCaps{2}) Drawing external negatives, (\RomanNumeralCaps{3}) Using the same encoder and MLP for one mapping instead of two, (\RomanNumeralCaps{4}) Adding cycle-consistency loss, and (\RomanNumeralCaps{5}) Removing the dual setting. We show the \underline{randomly} picked qualitative results in Figure~\ref{fig:supp3} and Figure~\ref{fig:supp4}. The qualitative results suggest that DCLGAN generates more realistic images than other variants, while each of our contribution has shown its efficiency.

\subsection{Additional Results}
We evaluated our proposed method and baselines among nine tasks. We choose the best four methods and show more qualitative results among all tasks except for Facade $\rightarrow$ Label. This is an extension of Figure 2 and Figure 3 in the main paper. Figure~\ref{fig:supp5} shows some qualitative results of Horse $\leftrightarrow$ Zebra, Figure~\ref{fig:supp6} shows the results of Cat $\leftrightarrow$ Dog. The results of CityScapes and Label $\rightarrow$ Facade are shown in Figure~\ref{fig:supp7}, Figure~\ref{fig:supp8} shows the results of Van Gogh $\rightarrow$ Photo and Orange $\rightarrow$ Apple. From Figure~\ref{fig:supp5}, we can observe that DCLGAN does not show perfect inference results in Horse $\leftrightarrow$ Zebra tasks. This is mainly due to the limitation of the dataset, where horse images are collected from ImageNet using the keyword wild horse. Although DCLGAN performs better than all other state-of-the-art methods among multiple challenging tasks, similar to most recent methods, it sometimes fails to distinguish the foreground and background.

Cat $\leftrightarrow$ Dog task requires geometry changes to match the distribution. As shown in Figure~\ref{fig:supp6}, DCLGAN performs the best in geometry changes and generates realistic cats/dogs with reasonable structure while CycleGAN~\cite{CycleGAN2017} fails to perform any geometry change. For texture changes, DCLGAN consistently outperforms all other methods on the whole. This is shown on the third row of Figure~\ref{fig:supp8} when CUT~\cite{park2020cut} fails to modify the color of whole oranges.

\section{Discussions}
\subsection{Similarity loss and mode collapse}
The degenerated solution for similarity loss is avoided by the constraints from other losses. The features sent to $H_{xr}, H_{xf}$ or $H_{yr}, H_{yf}$ are also different at each iteration, where the features do not represent patches with the same location.

Similarity loss prevents mode collapse, this is due to the mode collapse outputs not only lack diversities but also tend to be unrealistic. However, the diverse real images are always of good quality. Thus, when there is a potential mode collapse issue, the similarity loss increases, and behaves like a regularization term, to avoid mode collapse.

\subsection{External negatives}
We present the effect of drawing external negatives in section~\ref{ablation}, ablation \textbf{(\RomanNumeralCaps{2})}. We show that drawing external negatives in the same manner as SimCLR~\cite{chen2020simple} may have a positive influence on certain tasks quantitatively, but external negatives usually negatively affect qualitative results. This is shown in figure~\ref{fig:supp3}, where the generated pedestrians \& cars tend to be merged together. We agree with \cite{park2020cut} that internal negatives are more powerful than external negatives, however, we hypothesize that drawing external negatives in different ways may lead to different conclusions which can be investigated in the future.

\input{ figs/supp3}
\input{ figs/supp4}
\input{ figs/supp5}
\input{ figs/supp6}
\input{ figs/supp7}
\input{ figs/supp8}
\end{document}

%% file: figs/DCLGAN.tex
\begin{figure*}[htb]
    \centering
    \includegraphics[width=\textwidth,height = 3.2in] {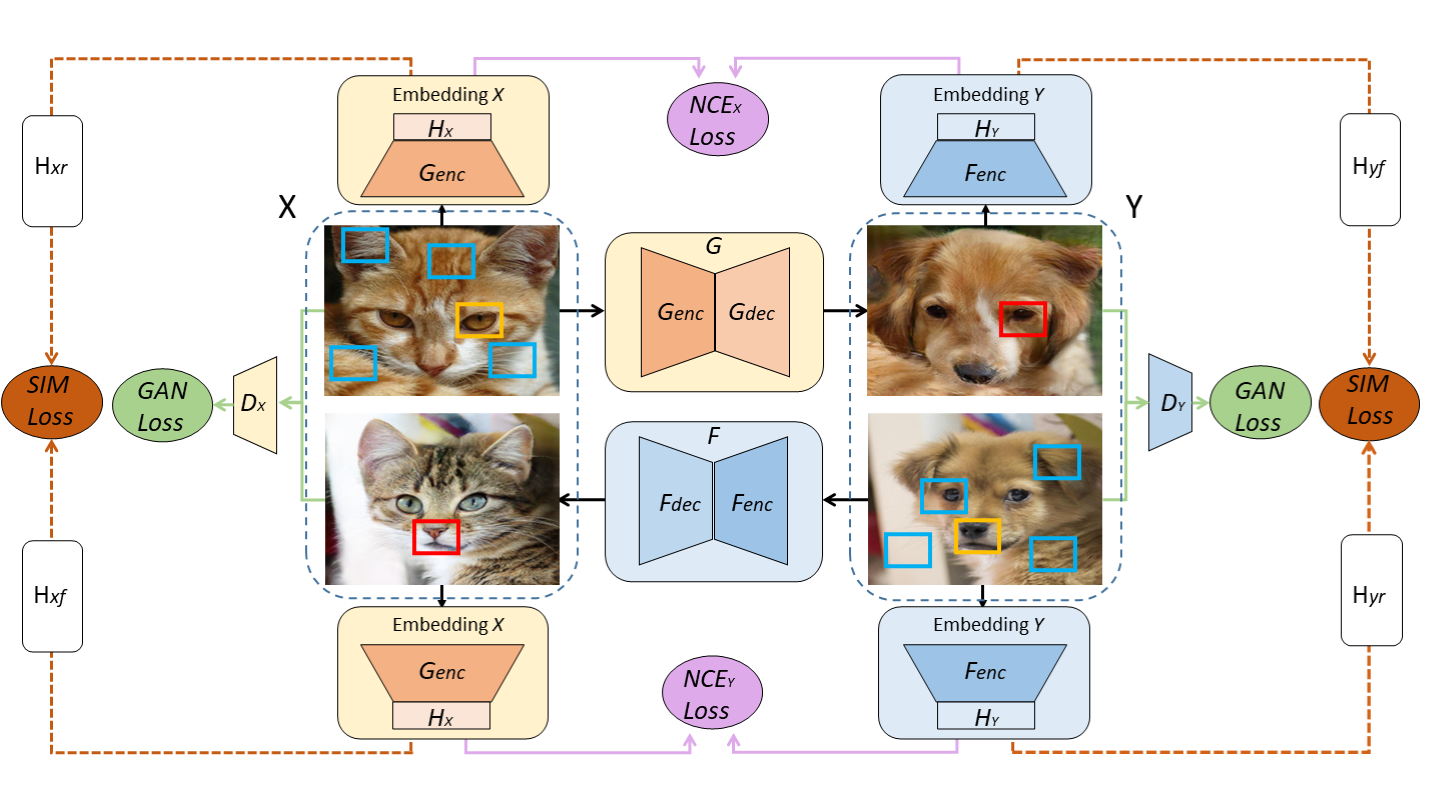}
     \caption{Overall architecture of DCLGAN: By dually learning two mappings $G: X\rightarrow Y$ and $F: Y\rightarrow X$, we successfully enable unpaired image-to-image translation without cycle-consistency. We define the encode half of $G$ and $F$ to be $G_{enc}, F_{enc}$. We use $G_{enc}$ and $H_{X}$ as embedding $X$ and $F_{enc}$ and $H_{Y}$ as embedding $Y$. We depict here the GAN loss (\textcolor{mygreen}{green line}), the patch-based multiplayer PatchNCE loss (\textcolor{mypurple}{purple line}). We omit the identical loss here. For variant SimDCL, we add the similarity loss between real images and fake images belonging to the same domain (\textcolor{mybrown}{dashed orange line}). PatchNCE loss helps the generated fake image \textcolor{red}{red} patch to be similar to its real input image \textcolor{myorange}{yellow} patch while dissimilar to other \textcolor{myblue}{blue} patches.}  
     \label{fig:DCLGAN}
\end{figure*}

%% file: figs/comparisonfull.tex
\begin{figure*}[!tb]
  \begin{minipage}[t]{0.087\linewidth} 
    \centering 
    \text{\small Input}
  \end{minipage} 
    \begin{minipage}[t]{0.087\linewidth} 
    \centering 
    \text{\small DCLGAN}
  \end{minipage} 
    \begin{minipage}[t]{0.087\linewidth} 
    \centering 
        \text{\small SimDCL}
  \end{minipage} 
    \begin{minipage}[t]{0.087\linewidth} 
    \centering 
        \text{\small CUT}
  \end{minipage} 
    \begin{minipage}[t]{0.087\linewidth} 
    \centering 
        \text{\small FastCUT}
  \end{minipage} 
    \begin{minipage}[t]{0.087\linewidth} 
    \centering 
            \text{\small CycleGAN}
  \end{minipage} 
    \begin{minipage}[t]{0.087\linewidth} 
    \centering 
            \text{\small MUNIT}
  \end{minipage} 
    \begin{minipage}[t]{0.087\linewidth} 
    \centering 
            \text{\small DRIT}
  \end{minipage} 
    \begin{minipage}[t]{0.087\linewidth} 
    \centering 
            \text{\small SelfDis}
  \end{minipage} 
    \begin{minipage}[t]{0.087\linewidth} 
    \centering 
            \text{\small Distance}
  \end{minipage} 
    \begin{minipage}[t]{0.087\linewidth} 
    \centering 
            \text{\small GCGAN}
  \end{minipage} 
  \\
  \begin{minipage}[t]{0.087\linewidth} 
    \centering 

    \includegraphics[width=0.57in, height=0.57in]{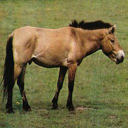}
  \end{minipage} 
    \begin{minipage}[t]{0.087\linewidth} 
    \centering 

    \includegraphics[width=0.57in, height=0.57in]{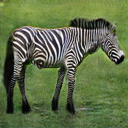}
  \end{minipage} 
      \begin{minipage}[t]{0.087\linewidth} 
    \centering 

    \includegraphics[width=0.57in, height=0.57in]{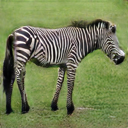}
  \end{minipage} 
    \begin{minipage}[t]{0.087\linewidth} 
    \centering 

    \includegraphics[width=0.57in, height=0.57in]{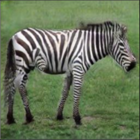}
  \end{minipage} 
    \begin{minipage}[t]{0.087\linewidth} 
    \centering 

    \includegraphics[width=0.57in, height=0.57in]{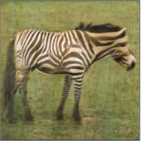}
  \end{minipage} 
    \begin{minipage}[t]{0.087\linewidth} 
    \centering 

    \includegraphics[width=0.57in, height=0.57in]{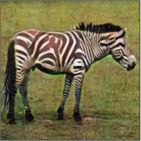}
  \end{minipage} 
    \begin{minipage}[t]{0.087\linewidth} 
    \centering 

    \includegraphics[width=0.57in, height=0.57in]{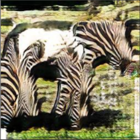}
  \end{minipage} 
    \begin{minipage}[t]{0.087\linewidth} 
    \centering 

    \includegraphics[width=0.57in, height=0.57in]{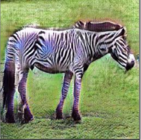}
  \end{minipage} 
    \begin{minipage}[t]{0.087\linewidth} 
    \centering 

    \includegraphics[width=0.57in, height=0.57in]{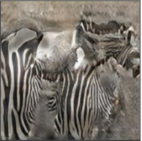}
  \end{minipage} 
    \begin{minipage}[t]{0.087\linewidth} 
    \centering 

    \includegraphics[width=0.57in, height=0.57in]{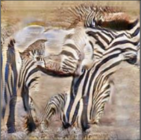}
  \end{minipage} 
    \begin{minipage}[t]{0.087\linewidth} 
    \centering 

    \includegraphics[width=0.57in, height=0.57in]{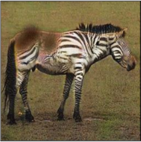}
  \end{minipage} 
  \\
      \begin{minipage}[t]{0.087\linewidth} 
    \centering 
    \includegraphics[width=0.57in, height=0.57in]{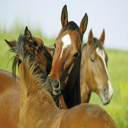}
  \end{minipage} 
    \begin{minipage}[t]{0.087\linewidth} 
    \centering 
    \includegraphics[width=0.57in, height=0.57in]{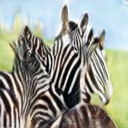}
  \end{minipage} 
      \begin{minipage}[t]{0.087\linewidth} 
    \centering 

    \includegraphics[width=0.57in, height=0.57in]{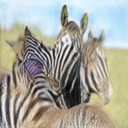}
  \end{minipage} 
    \begin{minipage}[t]{0.087\linewidth} 
    \centering 
    \includegraphics[width=0.57in, height=0.57in]{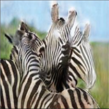}
  \end{minipage} 
    \begin{minipage}[t]{0.087\linewidth} 
    \centering 
    \includegraphics[width=0.57in, height=0.57in]{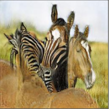}
  \end{minipage} 
    \begin{minipage}[t]{0.087\linewidth} 
    \centering 
    \includegraphics[width=0.57in, height=0.57in]{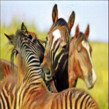}
  \end{minipage} 
    \begin{minipage}[t]{0.087\linewidth} 
    \centering 
    \includegraphics[width=0.57in]{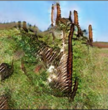}
  \end{minipage} 
    \begin{minipage}[t]{0.087\linewidth} 
    \centering 
    \includegraphics[width=0.57in, height=0.57in]{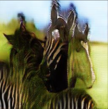}
  \end{minipage} 
    \begin{minipage}[t]{0.087\linewidth} 
    \centering 
    \includegraphics[width=0.57in, height=0.57in]{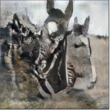}
  \end{minipage} 
    \begin{minipage}[t]{0.087\linewidth} 
    \centering 
    \includegraphics[width=0.57in, height=0.57in]{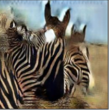}
  \end{minipage} 
    \begin{minipage}[t]{0.087\linewidth} 
    \centering 
    \includegraphics[width=0.57in, height=0.57in]{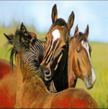}
  \end{minipage} 
  \\
        \begin{minipage}[t]{0.087\linewidth} 
    \centering 
    \includegraphics[width=0.57in, height=0.57in]{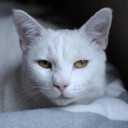}
  \end{minipage} 
    \begin{minipage}[t]{0.087\linewidth} 
    \centering 
    \includegraphics[width=0.57in, height=0.57in]{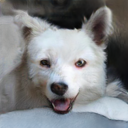}
  \end{minipage} 
      \begin{minipage}[t]{0.087\linewidth} 
    \centering 
    \includegraphics[width=0.57in, height=0.57in]{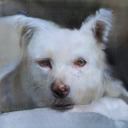}
  \end{minipage} 
    \begin{minipage}[t]{0.087\linewidth} 
    \centering 
    \includegraphics[width=0.57in, height=0.57in]{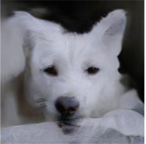}
  \end{minipage} 
    \begin{minipage}[t]{0.087\linewidth} 
    \centering 
    \includegraphics[width=0.57in, height=0.57in]{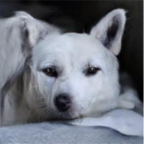}
  \end{minipage} 
    \begin{minipage}[t]{0.087\linewidth} 
    \centering 
    \includegraphics[width=0.57in, height=0.57in]{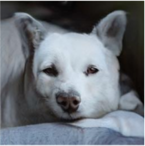}
  \end{minipage} 
    \begin{minipage}[t]{0.087\linewidth} 
    \centering 
    \includegraphics[width=0.57in, height=0.57in]{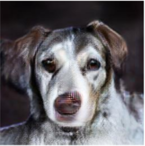}
  \end{minipage} 
    \begin{minipage}[t]{0.087\linewidth} 
    \centering 
    \includegraphics[width=0.57in, height=0.57in]{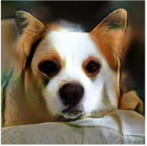}
  \end{minipage} 
    \begin{minipage}[t]{0.087\linewidth} 
    \centering 
    \includegraphics[width=0.57in, height=0.57in]{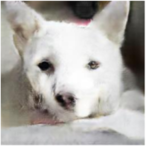}
  \end{minipage} 
    \begin{minipage}[t]{0.087\linewidth} 
    \centering 
    \includegraphics[width=0.57in, height=0.57in]{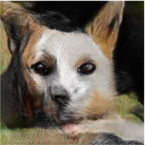}
  \end{minipage} 
    \begin{minipage}[t]{0.087\linewidth} 
    \centering 
    \includegraphics[width=0.57in, height=0.57in]{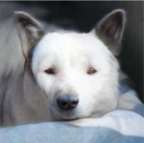}
  \end{minipage} 
        \begin{minipage}[t]{0.087\linewidth} 
    \centering 
    \includegraphics[width=0.57in, height=0.57in]{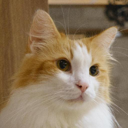}
  \end{minipage} 
    \begin{minipage}[t]{0.087\linewidth} 
    \centering 
    \includegraphics[width=0.57in, height=0.57in]{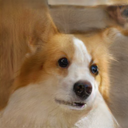}
  \end{minipage} 
      \begin{minipage}[t]{0.087\linewidth} 
    \centering 

    \includegraphics[width=0.57in, height=0.57in]{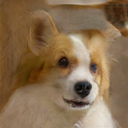}
  \end{minipage} 
    \begin{minipage}[t]{0.087\linewidth} 
    \centering 
    \includegraphics[width=0.57in, height=0.57in]{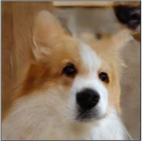}
  \end{minipage} 
    \begin{minipage}[t]{0.087\linewidth} 
    \centering 
    \includegraphics[width=0.57in, height=0.57in]{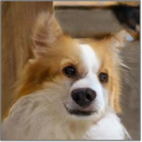}
  \end{minipage} 
    \begin{minipage}[t]{0.087\linewidth} 
    \centering 
    \includegraphics[width=0.57in, height=0.57in]{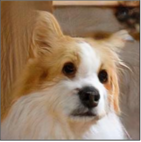}
  \end{minipage} 
    \begin{minipage}[t]{0.087\linewidth} 
    \centering 
    \includegraphics[width=0.57in, height=0.57in]{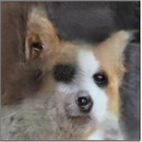}
  \end{minipage} 
    \begin{minipage}[t]{0.087\linewidth} 
    \centering 
    \includegraphics[width=0.57in, height=0.57in]{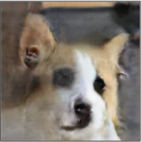}
  \end{minipage} 
    \begin{minipage}[t]{0.087\linewidth} 
    \centering 
    \includegraphics[width=0.57in, height=0.57in]{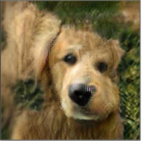}
  \end{minipage} 
    \begin{minipage}[t]{0.087\linewidth} 
    \centering 
    \includegraphics[width=0.57in, height=0.57in]{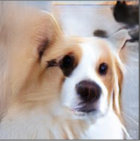}
  \end{minipage} 
    \begin{minipage}[t]{0.087\linewidth} 
    \centering 
    \includegraphics[width=0.57in, height=0.57in]{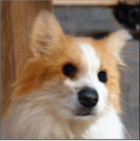}
  \end{minipage} 
  \\
        \begin{minipage}[t]{0.087\linewidth} 
    \centering 
    \includegraphics[width=0.57in, height=0.57in]{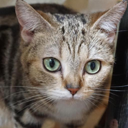}
  \end{minipage} 
    \begin{minipage}[t]{0.087\linewidth} 
    \centering 
    \includegraphics[width=0.57in, height=0.57in]{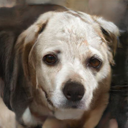}
  \end{minipage} 
      \begin{minipage}[t]{0.087\linewidth} 
    \centering 

    \includegraphics[width=0.57in, height=0.57in]{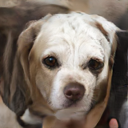}
  \end{minipage} 
    \begin{minipage}[t]{0.087\linewidth} 
    \centering 
    \includegraphics[width=0.57in, height=0.57in]{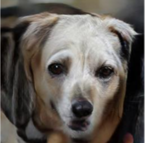}
  \end{minipage} 
    \begin{minipage}[t]{0.087\linewidth} 
    \centering 
    \includegraphics[width=0.57in, height=0.57in]{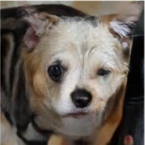}
  \end{minipage} 
    \begin{minipage}[t]{0.087\linewidth} 
    \centering 
    \includegraphics[width=0.57in, height=0.57in]{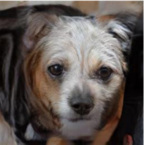}
  \end{minipage} 
    \begin{minipage}[t]{0.087\linewidth} 
    \centering 
    \includegraphics[width=0.57in, height=0.57in]{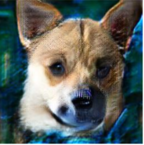}
  \end{minipage} 
    \begin{minipage}[t]{0.087\linewidth} 
    \centering 
    \includegraphics[width=0.57in, height=0.57in]{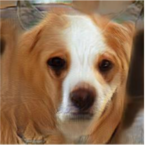}
  \end{minipage} 
    \begin{minipage}[t]{0.087\linewidth} 
    \centering 
    \includegraphics[width=0.57in, height=0.57in]{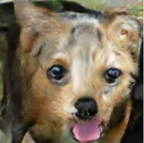}
  \end{minipage} 
    \begin{minipage}[t]{0.087\linewidth} 
    \centering 
    \includegraphics[width=0.57in, height=0.57in]{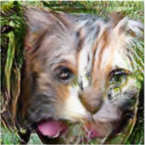}
  \end{minipage} 
    \begin{minipage}[t]{0.087\linewidth} 
    \centering 
    \includegraphics[width=0.57in, height=0.57in]{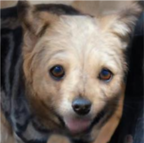}
  \end{minipage} 
  \\
        \begin{minipage}[t]{0.087\linewidth} 
    \centering 
    \includegraphics[width=0.57in, height=0.28in]{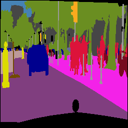}
  \end{minipage} 
    \begin{minipage}[t]{0.087\linewidth} 
    \centering 
    \includegraphics[width=0.57in,height=0.28in]{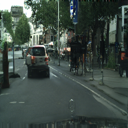}
  \end{minipage} 
      \begin{minipage}[t]{0.087\linewidth} 
    \centering 

    \includegraphics[width=0.57in, height=0.28in]{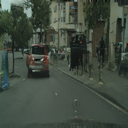}
  \end{minipage} 
    \begin{minipage}[t]{0.087\linewidth} 
    \centering 
    \includegraphics[width=0.57in,height=0.28in]{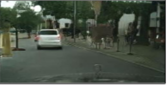}
  \end{minipage} 
    \begin{minipage}[t]{0.087\linewidth} 
    \centering 
    \includegraphics[width=0.57in,height=0.28in]{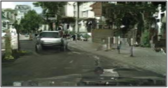}
  \end{minipage} 
    \begin{minipage}[t]{0.087\linewidth} 
    \centering 
    \includegraphics[width=0.57in,height=0.28in]{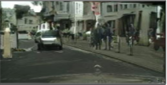}
  \end{minipage} 
    \begin{minipage}[t]{0.087\linewidth} 
    \centering 
    \includegraphics[width=0.57in,height=0.28in]{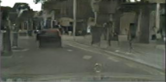}
  \end{minipage} 
    \begin{minipage}[t]{0.087\linewidth} 
    \centering 
    \includegraphics[width=0.57in,height=0.28in]{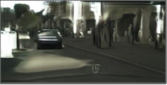}
  \end{minipage} 
    \begin{minipage}[t]{0.087\linewidth} 
    \centering 
    \includegraphics[width=0.57in,height=0.28in]{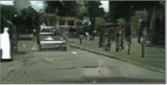}
  \end{minipage} 
    \begin{minipage}[t]{0.087\linewidth} 
    \centering 
    \includegraphics[width=0.57in,height=0.28in]{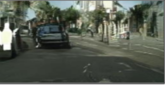}
  \end{minipage} 
    \begin{minipage}[t]{0.087\linewidth} 
    \centering 
    \includegraphics[width=0.57in,height=0.28in]{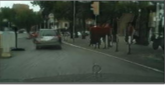}
  \end{minipage} 
  \\
        \begin{minipage}[t]{0.087\linewidth} 
    \centering 
    \includegraphics[width=0.57in, height=0.28in]{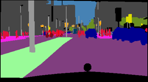}
  \end{minipage} 
    \begin{minipage}[t]{0.087\linewidth} 
    \centering 
    \includegraphics[width=0.57in,height=0.28in]{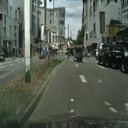}
  \end{minipage} 
      \begin{minipage}[t]{0.087\linewidth} 
    \centering 

    \includegraphics[width=0.57in, height=0.28in]{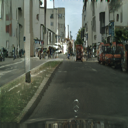}
  \end{minipage} 
    \begin{minipage}[t]{0.087\linewidth} 
    \centering 
    \includegraphics[width=0.57in,height=0.28in]{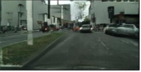}
  \end{minipage} 
    \begin{minipage}[t]{0.087\linewidth} 
    \centering 
    \includegraphics[width=0.57in,height=0.28in]{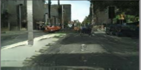}
  \end{minipage} 
    \begin{minipage}[t]{0.087\linewidth} 
    \centering 
    \includegraphics[width=0.57in,height=0.28in]{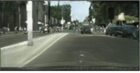}
  \end{minipage} 
    \begin{minipage}[t]{0.087\linewidth} 
    \centering 
    \includegraphics[width=0.57in,height=0.28in]{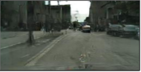}
  \end{minipage} 
    \begin{minipage}[t]{0.087\linewidth} 
    \centering 
    \includegraphics[width=0.57in,height=0.28in]{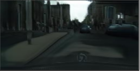}
  \end{minipage} 
    \begin{minipage}[t]{0.087\linewidth} 
    \centering 
    \includegraphics[width=0.57in,height=0.28in]{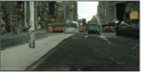}
  \end{minipage} 
    \begin{minipage}[t]{0.087\linewidth} 
    \centering 
    \includegraphics[width=0.57in,height=0.28in]{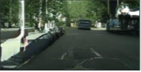}
  \end{minipage} 
    \begin{minipage}[t]{0.087\linewidth} 
    \centering 
    \includegraphics[width=0.57in,height=0.28in]{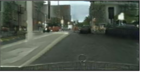}
  \end{minipage} 
\\
  \begin{minipage}[t]{0.087\linewidth} 
    \centering 
    \includegraphics[width=0.57in,height=0.57in]{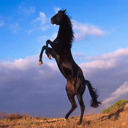}
  \end{minipage} 
    \begin{minipage}[t]{0.087\linewidth} 
    \centering 
    \includegraphics[width=0.57in,height=0.57in]{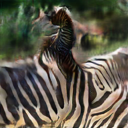}
  \end{minipage} 
      \begin{minipage}[t]{0.087\linewidth} 
    \centering 

    \includegraphics[width=0.57in, height=0.57in]{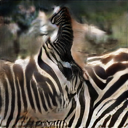}
  \end{minipage} 
    \begin{minipage}[t]{0.087\linewidth} 
    \centering 
    \includegraphics[width=0.57in,height=0.57in]{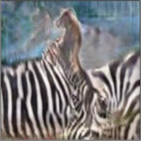}
  \end{minipage} 
    \begin{minipage}[t]{0.087\linewidth} 
    \centering 
    \includegraphics[width=0.57in]{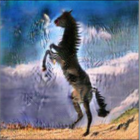}
  \end{minipage} 
    \begin{minipage}[t]{0.087\linewidth} 
    \centering 
    \includegraphics[width=0.57in,height=0.57in]{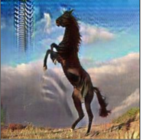}
  \end{minipage} 
    \begin{minipage}[t]{0.087\linewidth} 
    \centering 
    \includegraphics[width=0.57in,height=0.57in]{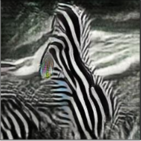}
  \end{minipage} 
    \begin{minipage}[t]{0.087\linewidth} 
    \centering 
    \includegraphics[width=0.57in,height=0.57in]{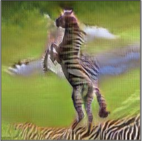}
  \end{minipage} 
    \begin{minipage}[t]{0.087\linewidth} 
    \centering 
    \includegraphics[width=0.57in,height=0.57in]{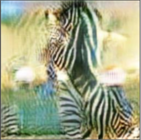}
  \end{minipage} 
    \begin{minipage}[t]{0.087\linewidth} 
    \centering 
    \includegraphics[width=0.57in,height=0.57in]{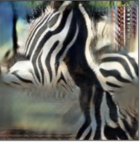}
  \end{minipage} 
    \begin{minipage}[t]{0.087\linewidth} 
    \centering 
    \includegraphics[width=0.57in,height=0.57in]{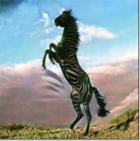}
  \end{minipage} 
  \caption{Comparison to all baselines on the Horse$\rightarrow$Zebra, Cat$\rightarrow$Dog, and CityScapes tasks. DCLGAN and SimDCL show visual satisfactory results. The last row is a failure case, our methods are unable to identify unusual pose and rare background. They fail to distinguish foreground and background, adding zebra textures to the cloud.}
  \label{fig:resultall}
\end{figure*}

%% file: tables/comparision.tex
\begin{table}[!htbp]
  \centering
  \fontsize{6.63}{3}\selectfont
    \begin{tabular}{cccccc}
    \toprule
    \multicolumn{2}{c}{ \textbf{\quad \quad \quad \quad \quad \quad \quad \quad CityScapes}}& \multicolumn{1}{c}{\textbf{Cat$\rightarrow$ Dog}}& \multicolumn{2}{c}{\textbf{Horse $\rightarrow$ Zebra}} & \multicolumn{1}{c}{\textbf{Overall}}  \cr
    \cmidrule(lr){2-2} \cmidrule(lr){3-3} \cmidrule(lr){4-5} 
    \textbf{Method}&\textbf{FID$\downarrow$}&\textbf{FID$\downarrow$}&\textbf{FID$\downarrow$}&\textbf{sec/iter$\downarrow$}&\textbf{Ranking} \cr
    \midrule
    CycleGAN~\cite{CycleGAN2017}&68.6&85.9&66.8&0.40&4\cr
    MUNIT~\cite{huang2018multimodal}&91.4&104.4&133.8&0.39&9\cr
    DRIT~\cite{kim2017disco}&155.3&123.4&140.0&0.70&10\cr
    Distance~\cite{distance}&85.8&155.3&72.0&\textbf{0.15}&6\cr
    SelfDistance~\cite{distance}&78.8&144.4&80.8& \underline{0.16}&6\cr
    GCGAN~\cite{geometry2019}&105.2&96.6&86.7&0.62&8\cr
    CUT~\cite{park2020cut}&56.4&76.2& \underline{45.5}&0.24&3\cr
    FastCUT~\cite{park2020cut}&68.8&94.0&73.4&\textbf{0.15}&5\cr
    \cmidrule(lr){1-6}
    DCLGAN (ours)&\textbf{49.4}&\textbf{60.7}&\textbf{43.2}&0.41&\textbf{1}\cr
    SimDCL (ours)& \underline{51.3}& \underline{65.5}&47.1&0.47& \underline{2}\cr
    \bottomrule
    \end{tabular}
     \caption{Comparison to all baselines on the Horse$\rightarrow$Zebra, Cat$\rightarrow$Dog, and CityScapes tasks. DCLGAN denotes our model without Similarity loss and SimDCL denotes our model with Similarity loss. We show FID~\cite{heusel2017gans} score for all tasks. The overall ranking is based on the FID score among all tasks. DCLGAN generates better images with acceptable speed, runs a bit slower than CycleGAN~\cite{CycleGAN2017}. Our variant SimDCL also shows competitive results.}
     \label{tab:performance_comparison}
\end{table}

%% file: figs/resultmore_double.tex
\begin{figure*}[htb]
  \begin{minipage}[t]{0.095\linewidth} 
    \centering 
    \text{\small Input}
  \end{minipage} 
    \begin{minipage}[t]{0.095\linewidth} 
    \centering 
    \text{\small DCLGAN}
  \end{minipage} 
    \begin{minipage}[t]{0.095\linewidth} 
    \centering 
      \text{\small SimDCL}
  \end{minipage} 
    \begin{minipage}[t]{0.095\linewidth} 
    \centering 
        \text{\small CUT}
  \end{minipage} 
    \begin{minipage}[t]{0.095\linewidth} 
    \centering 
        \text{\small CycleGAN}
  \end{minipage} 
    \begin{minipage}[t]{0.095\linewidth} 
    \centering 
    \text{\small Input}
  \end{minipage} 
    \begin{minipage}[t]{0.095\linewidth} 
    \centering 
    \text{\small DCLGAN}
  \end{minipage} 
    \begin{minipage}[t]{0.095\linewidth} 
    \centering 
      \text{\small SimDCL}
  \end{minipage} 
    \begin{minipage}[t]{0.095\linewidth} 
    \centering 
        \text{\small CUT}
  \end{minipage} 
    \begin{minipage}[t]{0.095\linewidth} 
    \centering 
        \text{\small CycleGAN}
  \end{minipage} 
  \\
  \begin{minipage}[t]{0.095\linewidth} 
    \centering 
    \includegraphics[width=0.62in, height=0.62in]{ 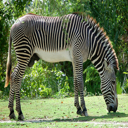}
  \end{minipage} 
    \begin{minipage}[t]{0.095\linewidth} 
    \centering 
    \includegraphics[width=0.62in, height=0.62in]{ 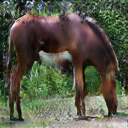}
  \end{minipage} 
      \begin{minipage}[t]{0.095\linewidth} 
    \centering 
    \includegraphics[width=0.62in, height=0.62in]{ 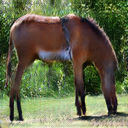}
  \end{minipage} 
    \begin{minipage}[t]{0.095\linewidth} 
    \centering 
    \includegraphics[width=0.62in, height=0.62in]{ 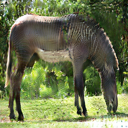}
  \end{minipage} 
    \begin{minipage}[t]{0.095\linewidth} 
    \centering 
    \includegraphics[width=0.62in, height=0.62in]{ 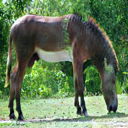}
  \end{minipage} 
  \begin{minipage}[t]{0.095\linewidth} 
    \centering 
    \includegraphics[width=0.62in, height=0.62in]{ 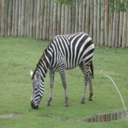}
  \end{minipage} 
    \begin{minipage}[t]{0.095\linewidth} 
    \centering 
    \includegraphics[width=0.62in, height=0.62in]{ 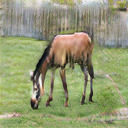}
  \end{minipage} 
      \begin{minipage}[t]{0.095\linewidth} 
    \centering 
    \includegraphics[width=0.62in, height=0.62in]{ 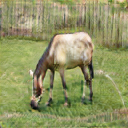}
  \end{minipage} 
    \begin{minipage}[t]{0.095\linewidth} 
    \centering 
    \includegraphics[width=0.62in, height=0.62in]{ 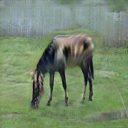}
  \end{minipage} 
    \begin{minipage}[t]{0.095\linewidth} 
    \centering 
    \includegraphics[width=0.62in, height=0.62in]{ 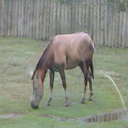}
  \end{minipage} 
\\
  \begin{minipage}[t]{0.095\linewidth} 
    \centering 
    \includegraphics[width=0.62in, height=0.62in]{ 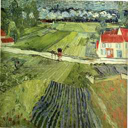}
  \end{minipage} 
    \begin{minipage}[t]{0.095\linewidth} 
    \centering 
    \includegraphics[width=0.62in, height=0.62in]{ 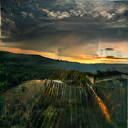}
  \end{minipage} 
      \begin{minipage}[t]{0.095\linewidth} 
    \centering 
    \includegraphics[width=0.62in, height=0.62in]{ 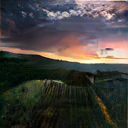}
  \end{minipage} 
    \begin{minipage}[t]{0.095\linewidth} 
    \centering 
    \includegraphics[width=0.62in, height=0.62in]{ 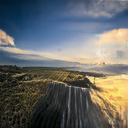}
  \end{minipage} 
    \begin{minipage}[t]{0.095\linewidth} 
    \centering 
    \includegraphics[width=0.62in, height=0.62in]{ 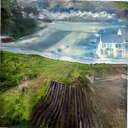}
  \end{minipage} 
  \begin{minipage}[t]{0.095\linewidth} 
    \centering 
    \includegraphics[width=0.62in, height=0.62in]{ 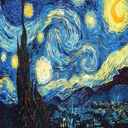}
  \end{minipage} 
    \begin{minipage}[t]{0.095\linewidth} 
    \centering 
    \includegraphics[width=0.62in, height=0.62in]{ 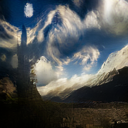}
  \end{minipage} 
      \begin{minipage}[t]{0.095\linewidth} 
    \centering 
    \includegraphics[width=0.62in, height=0.62in]{ 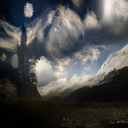}
  \end{minipage} 
    \begin{minipage}[t]{0.095\linewidth} 
    \centering 
    \includegraphics[width=0.62in, height=0.62in]{ 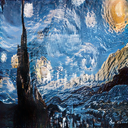}
  \end{minipage} 
    \begin{minipage}[t]{0.095\linewidth} 
    \centering 
    \includegraphics[width=0.62in, height=0.62in]{ 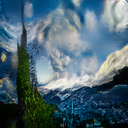}
  \end{minipage} 
\\
  \begin{minipage}[t]{0.095\linewidth} 
    \centering 
    \includegraphics[width=0.62in, height=0.62in]{ 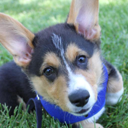}
  \end{minipage} 
    \begin{minipage}[t]{0.095\linewidth} 
    \centering 
    \includegraphics[width=0.62in, height=0.62in]{ 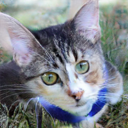}
  \end{minipage} 
      \begin{minipage}[t]{0.095\linewidth} 
    \centering 
    \includegraphics[width=0.62in, height=0.62in]{ 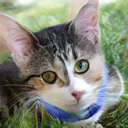}
  \end{minipage} 
    \begin{minipage}[t]{0.095\linewidth} 
    \centering 
    \includegraphics[width=0.62in, height=0.62in]{ 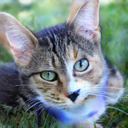}
  \end{minipage} 
    \begin{minipage}[t]{0.095\linewidth} 
    \centering 
    \includegraphics[width=0.62in, height=0.62in]{ 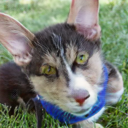}
  \end{minipage} 
  \begin{minipage}[t]{0.095\linewidth} 
    \centering 
    \includegraphics[width=0.62in, height=0.62in]{ 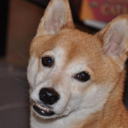}
  \end{minipage} 
    \begin{minipage}[t]{0.095\linewidth} 
    \centering 
    \includegraphics[width=0.62in, height=0.62in]{ 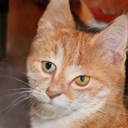}
  \end{minipage} 
      \begin{minipage}[t]{0.095\linewidth} 
    \centering 
    \includegraphics[width=0.62in, height=0.62in]{ 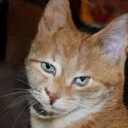}
  \end{minipage} 
    \begin{minipage}[t]{0.095\linewidth} 
    \centering 
    \includegraphics[width=0.62in, height=0.62in]{ 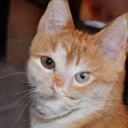}
  \end{minipage} 
    \begin{minipage}[t]{0.095\linewidth} 
    \centering 
    \includegraphics[width=0.62in, height=0.62in]{ 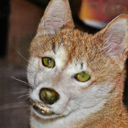}
  \end{minipage} 
\\
  \begin{minipage}[t]{0.095\linewidth} 
    \centering 
    \includegraphics[width=0.62in, height=0.62in]{ 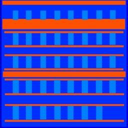}
  \end{minipage} 
    \begin{minipage}[t]{0.095\linewidth} 
    \centering 
    \includegraphics[width=0.62in, height=0.62in]{ 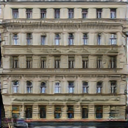}
  \end{minipage} 
      \begin{minipage}[t]{0.095\linewidth} 
    \centering 
    \includegraphics[width=0.62in, height=0.62in]{ 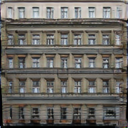}
  \end{minipage} 
    \begin{minipage}[t]{0.095\linewidth} 
    \centering 
    \includegraphics[width=0.62in, height=0.62in]{ 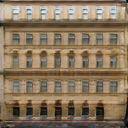}
  \end{minipage} 
    \begin{minipage}[t]{0.095\linewidth} 
    \centering 
    \includegraphics[width=0.62in, height=0.62in]{ 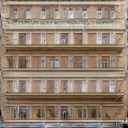}
  \end{minipage} 
  \begin{minipage}[t]{0.095\linewidth} 
    \centering 
    \includegraphics[width=0.62in, height=0.62in]{ 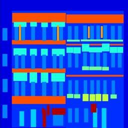}
  \end{minipage} 
    \begin{minipage}[t]{0.095\linewidth} 
    \centering 
    \includegraphics[width=0.62in, height=0.62in]{ 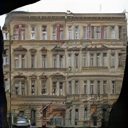}
  \end{minipage} 
      \begin{minipage}[t]{0.095\linewidth} 
    \centering 
    \includegraphics[width=0.62in, height=0.62in]{ 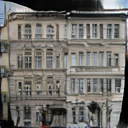}
  \end{minipage} 
    \begin{minipage}[t]{0.095\linewidth} 
    \centering 
    \includegraphics[width=0.62in, height=0.62in]{ 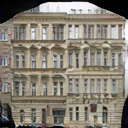}
  \end{minipage} 
    \begin{minipage}[t]{0.095\linewidth} 
    \centering 
    \includegraphics[width=0.62in, height=0.62in]{ 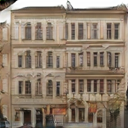}
  \end{minipage} 
\\
  \begin{minipage}[t]{0.095\linewidth} 
    \centering 
    \includegraphics[width=0.62in, height=0.62in]{ 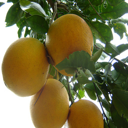}
  \end{minipage} 
    \begin{minipage}[t]{0.095\linewidth} 
    \centering 
    \includegraphics[width=0.62in, height=0.62in]{ 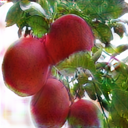}
  \end{minipage} 
      \begin{minipage}[t]{0.095\linewidth} 
    \centering 
    \includegraphics[width=0.62in, height=0.62in]{ 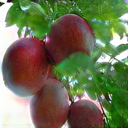}
  \end{minipage} 
    \begin{minipage}[t]{0.095\linewidth} 
    \centering 
    \includegraphics[width=0.62in, height=0.62in]{ 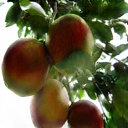}
  \end{minipage} 
    \begin{minipage}[t]{0.095\linewidth} 
    \centering 
    \includegraphics[width=0.62in, height=0.62in]{ 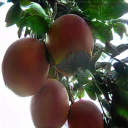}
  \end{minipage} 
  \begin{minipage}[t]{0.095\linewidth} 
    \centering 
    \includegraphics[width=0.62in, height=0.62in]{ 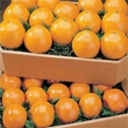}
  \end{minipage} 
    \begin{minipage}[t]{0.095\linewidth} 
    \centering 
    \includegraphics[width=0.62in, height=0.62in]{ 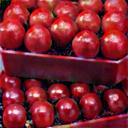}
  \end{minipage} 
      \begin{minipage}[t]{0.095\linewidth} 
    \centering 
    \includegraphics[width=0.62in, height=0.62in]{ 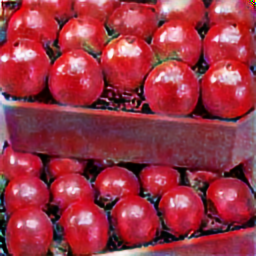}
  \end{minipage} 
    \begin{minipage}[t]{0.095\linewidth} 
    \centering 
    \includegraphics[width=0.62in, height=0.62in]{ 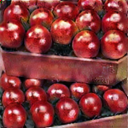}
  \end{minipage} 
    \begin{minipage}[t]{0.095\linewidth} 
    \centering 
    \includegraphics[width=0.62in, height=0.622in]{ 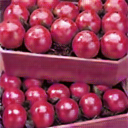}
  \end{minipage} 
  \\
  \begin{minipage}[t]{0.095\linewidth} 
    \centering 
    \includegraphics[width=0.622in, height=0.62in]{ 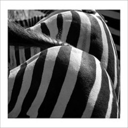}
  \end{minipage} 
    \begin{minipage}[t]{0.095\linewidth} 
    \centering 
    \includegraphics[width=0.62in, height=0.62in]{ 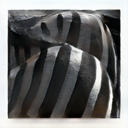}
  \end{minipage} 
      \begin{minipage}[t]{0.095\linewidth} 
    \centering 
    \includegraphics[width=0.62in, height=0.62in]{ 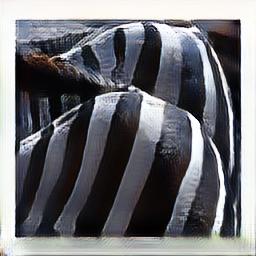}
  \end{minipage} 
    \begin{minipage}[t]{0.095\linewidth} 
    \centering 
    \includegraphics[width=0.62in, height=0.62in]{ 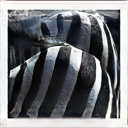}
  \end{minipage} 
    \begin{minipage}[t]{0.095\linewidth} 
    \centering 
    \includegraphics[width=0.62in, height=0.62in]{ 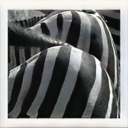}
  \end{minipage} 
  \begin{minipage}[t]{0.095\linewidth} 
    \centering 
    \includegraphics[width=0.62in, height=0.62in]{ 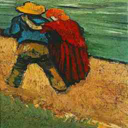}
  \end{minipage} 
    \begin{minipage}[t]{0.095\linewidth} 
    \centering 
    \includegraphics[width=0.62in, height=0.62in]{ 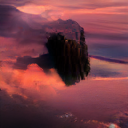}
  \end{minipage} 
      \begin{minipage}[t]{0.095\linewidth} 
    \centering 
    \includegraphics[width=0.62in, height=0.62in]{ 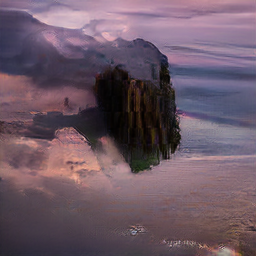}
  \end{minipage} 
    \begin{minipage}[t]{0.095\linewidth} 
    \centering 
    \includegraphics[width=0.62in, height=0.62in]{ 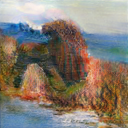}
  \end{minipage} 
    \begin{minipage}[t]{0.095\linewidth} 
    \centering 
    \includegraphics[width=0.62in, height=0.62in]{ 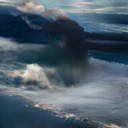}
  \end{minipage} 
  \caption{Comparison between the best four methods on five more tasks including Zebra $\rightarrow$ Horse, Van Gogh $\rightarrow$ Photo, Dog $\rightarrow$ Cat, Label $\rightarrow$ Facade, and Orange $\rightarrow$ Apple. We randomly pick two samples for each task. Our DCLGAN performs both geometry changes and texture changes. The last row show two typical failure cases, the first one fails to translate the image since input is only a small part of zebra, the last one fails to keep the structure of humans, translating them to Yosemite.}
  \label{fig:second}
\end{figure*}

%% file: tables/comparisionmore.txt
\begin{table*}[!htbp]
  \centering
  \fontsize{8}{4}\selectfont
    \begin{tabular}{cccccccc}
    \toprule
    \multicolumn{2}{c}{\textbf{\quad \quad \quad \quad \quad \quad \quad \quad  Zebra$\rightarrow$ Horse}}& \multicolumn{1}{c}{\textbf{ Van Gogh $\rightarrow$ Photo}}& \multicolumn{2}{c}{\textbf{Dog $\rightarrow$ Cat}} & \multicolumn{1}{c}{\textbf{Label $\rightarrow$ Facade}} &
\multicolumn{1}{c}{\textbf{Orange $\rightarrow$ Apple}} &
\multicolumn{1}{c}{\textbf{Model}} \cr

    \cmidrule(lr){2-2} \cmidrule(lr){3-3} \cmidrule(lr){4-5}  \cmidrule(lr){6-6} \cmidrule(lr){7-7} 
    \textbf{Method}&\textbf{FID$\downarrow$}&\textbf{FID$\downarrow$}&\textbf{Overall Runtime$\downarrow$}
&\textbf{FID$\downarrow$}&\textbf{FID$\downarrow$}&\textbf{FID$\downarrow$} &\textbf{Parameters}\cr

    \midrule
    CycleGAN~\cite{CycleGAN2017}&154.3&103.0&106hr&107.7&127.5 &\textbf{117.7}& 28.286M\cr
    CUT~\cite{park2020cut}&170.5&96.9&125hr&26.8& \underline{119.7} &127.0& 14.406M\cr
    DCLGAN (ours)&\textbf{139.5}& \underline{93.7}&108hr&\textbf{22.2}&\textbf{119.2} & \underline{124.9}&28.812M\cr
    SimDCL (ours)& \underline{152.5}&\textbf{93.5}&124hr& \underline{22.8}&132.3&134.4& 28.852M\cr
    \bottomrule
    \end{tabular}
      \caption{Comparison between the best four methods on Zebra $\rightarrow$ Horse, Van Gogh $\rightarrow$ Photo, Dog $\rightarrow$ Cat, and Label $\rightarrow$ Facade tasks. DCLGAN still outperforms other methods in most tasks. The overall runtimes are provided for Dog $\rightarrow$ Cat task, in hours. Note CUT is trained for 400 epochs while the rest for 200 epochs only. The overall ranking circumstances among methods compared in here are identical to the first comparison (Table~\ref{tab:performance_comparison}) except for a tie with CycleGAN~\cite{CycleGAN2017} and CUT~\cite{park2020cut}.}
  \label{tab:performance_more}
\end{table*}

%% file: tables/3compare.tex
\begin{table}[!htbp]
  \centering
  \fontsize{9}{3}\selectfont
    \begin{tabular}{cccc}
    \toprule
    \multicolumn{4}{c}{\textbf{\quad \quad \quad \quad \quad \quad \quad \quad \quad CityScapes}}\cr
    \cmidrule(lr){2-4}
    \textbf{Method}&\textbf{pixAcc$\uparrow$}&\textbf{classAcc$\uparrow$}&\textbf{IoU$\uparrow$} \cr
    \midrule
    DCLGAN(ours)&\textbf{0.74}&0.22&0.17\cr
    \cmidrule(lr){1-4}
    Pix2Pix~\cite{isola2017image} &0.66&\underline{0.23}&0.17\cr
    CRN~\cite{chen2017photographic} &0.69&0.21&\textbf{0.20}\cr 
    DRPAN~\cite{wang2018discriminative} &\underline{0.73}&\textbf{0.24}&\underline{0.19}\cr 
    \cmidrule(lr){1-4}    
    Ground Truth&0.80&0.26&0.21\cr 
    \bottomrule
    \end{tabular}
     \caption{Comparison between unsupervised DCLGAN and supervised Pix2Pix~\cite{isola2017image}, CRN~\cite{chen2017photographic}, DRPAN~\cite{wang2018discriminative} on CityScapes dataset. We follow the setting of Pix2Pix~\cite{isola2017image} to compute the FCN~\cite{long2015fully} score. DCLGAN outperforms supervised methods in pixAcc, suggesting the gap between unsupervised methods and supervised methods is closing.}
     \label{tab:3compare}
\end{table}

%% file: figs/facade.tex
\begin{figure}[htbp]
  \begin{minipage}[t]{0.155\linewidth} 
    \centering 
    \text{\small Input}
  \end{minipage} 
    \begin{minipage}[t]{0.155\linewidth} 
    \centering 
    \text{\small DCLGAN}
  \end{minipage} 
    \begin{minipage}[t]{0.155\linewidth} 
    \centering 
      \text{\small SimDCL}
  \end{minipage} 
    \begin{minipage}[t]{0.155\linewidth} 
    \centering 
        \text{\small CUT}
  \end{minipage} 
    \begin{minipage}[t]{0.155\linewidth} 
    \centering 
        \text{\small CycleGAN}
  \end{minipage} 
    \begin{minipage}[t]{0.155\linewidth} 
    \centering 
            \text{\small Label}
  \end{minipage} 
  \\
  \begin{minipage}[t]{0.155\linewidth} 
    \centering 
    \includegraphics[width=0.51in, height=0.51in]{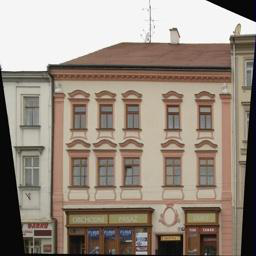}
  \end{minipage} 
    \begin{minipage}[t]{0.155\linewidth} 
    \centering 

    \includegraphics[width=0.51in, height=0.51in]{ 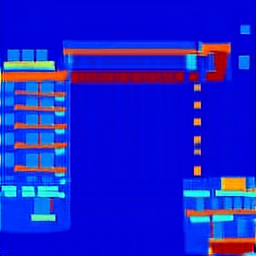}
  \end{minipage} 
      \begin{minipage}[t]{0.155\linewidth} 
    \centering 

    \includegraphics[width=0.51in, height=0.51in]{ 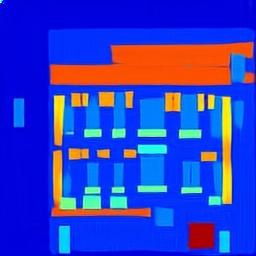}
  \end{minipage} 
    \begin{minipage}[t]{0.155\linewidth} 
    \centering 

    \includegraphics[width=0.51in, height=0.51in]{ 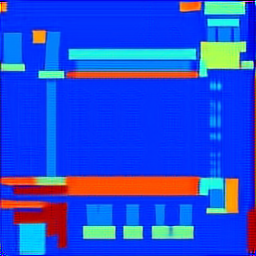}
  \end{minipage} 
    \begin{minipage}[t]{0.155\linewidth} 
    \centering 

    \includegraphics[width=0.51in, height=0.51in]{ 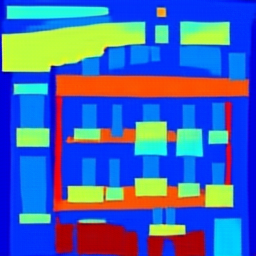}
  \end{minipage} 
    \begin{minipage}[t]{0.155\linewidth} 
    \centering 
    \includegraphics[width=0.51in, height=0.51in]{ 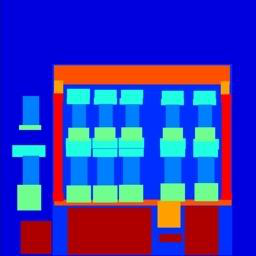}
  \end{minipage} 
  \\
    \begin{minipage}[t]{0.155\linewidth} 
    \centering 
    \includegraphics[width=0.51in, height=0.51in]{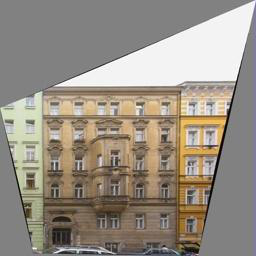}
  \end{minipage} 
    \begin{minipage}[t]{0.155\linewidth} 
    \centering 

    \includegraphics[width=0.51in, height=0.51in]{ 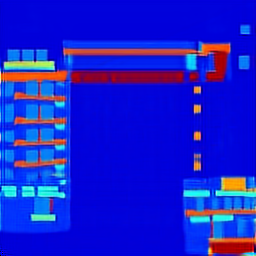}
  \end{minipage} 
      \begin{minipage}[t]{0.155\linewidth} 
    \centering 

    \includegraphics[width=0.51in, height=0.51in]{ 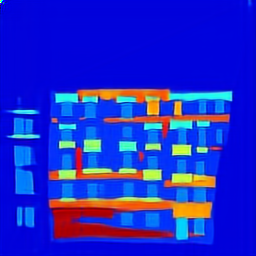}
  \end{minipage} 
    \begin{minipage}[t]{0.155\linewidth} 
    \centering 

    \includegraphics[width=0.51in, height=0.51in]{ 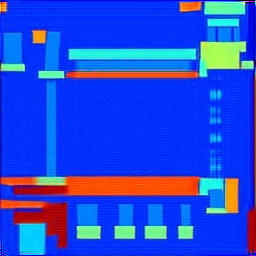}
  \end{minipage} 
    \begin{minipage}[t]{0.155\linewidth} 
    \centering 

    \includegraphics[width=0.51in, height=0.51in]{ 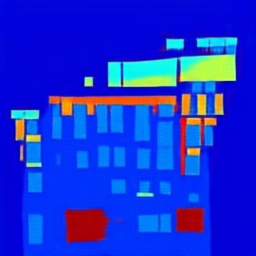}
  \end{minipage} 
    \begin{minipage}[t]{0.155\linewidth} 
    \centering 
    \includegraphics[width=0.51in, height=0.51in]{ 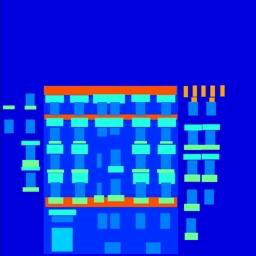}
  \end{minipage} 
    \\
    \begin{minipage}[t]{0.155\linewidth} 
    \centering 
    \includegraphics[width=0.51in, height=0.51in]{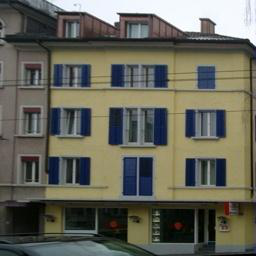}
  \end{minipage} 
    \begin{minipage}[t]{0.155\linewidth} 
    \centering 

    \includegraphics[width=0.51in, height=0.51in]{ 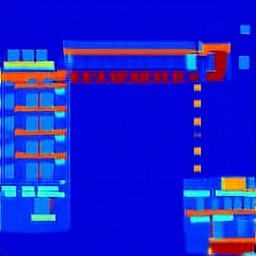}
  \end{minipage} 
      \begin{minipage}[t]{0.155\linewidth} 
    \centering 

    \includegraphics[width=0.51in, height=0.51in]{ 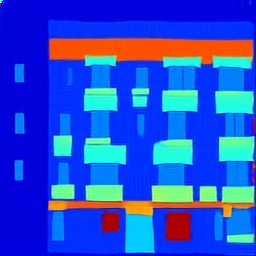}
  \end{minipage} 
    \begin{minipage}[t]{0.155\linewidth} 
    \centering 

    \includegraphics[width=0.51in, height=0.51in]{ 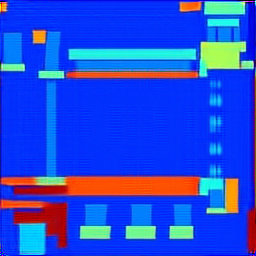}
  \end{minipage} 
    \begin{minipage}[t]{0.155\linewidth} 
    \centering 

    \includegraphics[width=0.51in, height=0.51in]{ 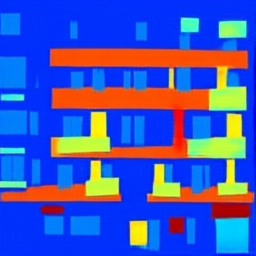}
  \end{minipage} 
    \begin{minipage}[t]{0.155\linewidth} 
    \centering 
    \includegraphics[width=0.51in, height=0.51in]{ 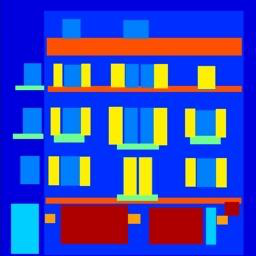}
  \end{minipage} 
  \caption{Comparison between the best four methods on Facade $\rightarrow$ Label task. Methods based on mutual information maximization suffer from mode collapse in this task. We address this by introducing SimDCL to prevent mode collapse. SimDCL also capture more correspondence between facade and label than CycleGAN~\cite{CycleGAN2017}.}
  \label{fig:facade}
\end{figure}

%% file: tables/ablation.tex
\begin{table}[!htp]
  \centering
  \fontsize{8.5}{10}\selectfont
    \begin{tabular}{cccc}
    \toprule
    \multicolumn{2}{c}{\textbf{\quad \quad \quad \quad \quad Horse $\rightarrow$ Zebra}}& \multicolumn{1}{c}{\textbf{Zebra $\rightarrow$ Horse}}&
    \multicolumn{1}{c}{\textbf{CityScapes}}\cr
    \cmidrule(lr){2-2} \cmidrule(lr){3-3} \cmidrule(lr){4-4}
    \textbf{Ablation}
    &\textbf{FID$\downarrow$ }
    &\textbf{FID$\downarrow$} &\textbf{FID$\downarrow$}\cr
    \midrule
    \RomanNumeralCaps{1}&49.7&156.7& 50.3\cr   
    \RomanNumeralCaps{2}&\textbf{41.7}&149.2&\underline{49.6}\cr   
    \RomanNumeralCaps{3}& 44.0&153.4&52.2\cr   
    \RomanNumeralCaps{4}&44.6& \underline{140.6}&55.4\cr   
    \RomanNumeralCaps{5}&47.0&151.3&91.5\cr
    \cmidrule(lr){1-4}
    DCLGAN&\underline{43.2}&\textbf{139.5}&\textbf{49.4}\cr
    \bottomrule
    \end{tabular}
    \caption{Quantitative results for ablations.}
    \label{tab:ablation}
\end{table}

%% file: figs/supp3.tex
\begin{figure*}[!tb]
  \begin{minipage}[t]{0.13\linewidth} 
    \centering 
    \text{\small Input}
  \end{minipage} 
    \begin{minipage}[t]{0.13\linewidth} 
    \centering 
        \text{\RomanNumeralCaps{1}}
  \end{minipage} 
    \begin{minipage}[t]{0.13\linewidth} 
    \centering 
        \text{\RomanNumeralCaps{2}}
  \end{minipage} 
    \begin{minipage}[t]{0.13\linewidth} 
    \centering 
        \text{\RomanNumeralCaps{3}}
  \end{minipage} 
    \begin{minipage}[t]{0.13\linewidth} 
    \centering 
            \text{\RomanNumeralCaps{4}}
  \end{minipage} 
    \begin{minipage}[t]{0.13\linewidth} 
    \centering 
            \text{\RomanNumeralCaps{5}}
  \end{minipage} 
      \begin{minipage}[t]{0.13\linewidth} 
    \centering 
    \text{\small DCLGAN}
  \end{minipage} 
  \\
  \begin{minipage}[t]{0.13\linewidth} 
    \centering 
    \includegraphics[width=0.9in, height=0.6in]{ 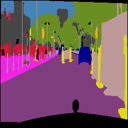}
  \end{minipage} 
      \begin{minipage}[t]{0.13\linewidth} 
    \centering 
    \includegraphics[width=0.9in, height=0.6in]{ 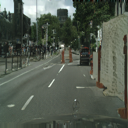}
  \end{minipage} 
    \begin{minipage}[t]{0.13\linewidth} 
    \centering 
    \includegraphics[width=0.9in, height=0.6in]{ 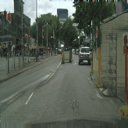}
  \end{minipage} 
    \begin{minipage}[t]{0.13\linewidth} 
    \centering 
    \includegraphics[width=0.9in, height=0.6in]{ 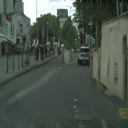}
  \end{minipage} 
    \begin{minipage}[t]{0.13\linewidth} 
    \centering 
    \includegraphics[width=0.9in, height=0.6in]{ 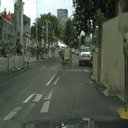}
  \end{minipage} 
    \begin{minipage}[t]{0.13\linewidth} 
    \centering 
    \includegraphics[width=0.9in, height=0.6in]{ 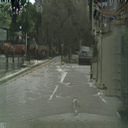}
  \end{minipage} 
      \begin{minipage}[t]{0.13\linewidth} 
    \centering 
    \includegraphics[width=0.9in, height=0.6in]{ 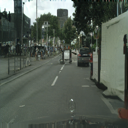}
  \end{minipage} 
\\
  \begin{minipage}[t]{0.13\linewidth} 
    \centering 
    \includegraphics[width=0.9in, height=0.6in]{ 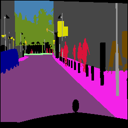}
  \end{minipage} 
      \begin{minipage}[t]{0.13\linewidth} 
    \centering 
    \includegraphics[width=0.9in, height=0.6in]{ 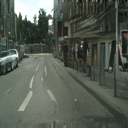}
  \end{minipage} 
    \begin{minipage}[t]{0.13\linewidth} 
    \centering 
    \includegraphics[width=0.9in, height=0.6in]{ 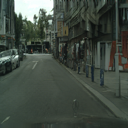}
  \end{minipage} 
    \begin{minipage}[t]{0.13\linewidth} 
    \centering 
    \includegraphics[width=0.9in, height=0.6in]{ 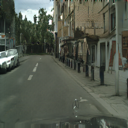}
  \end{minipage} 
    \begin{minipage}[t]{0.13\linewidth} 
    \centering 
    \includegraphics[width=0.9in, height=0.6in]{ 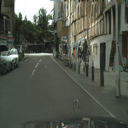}
  \end{minipage} 
    \begin{minipage}[t]{0.13\linewidth} 
    \centering 
    \includegraphics[width=0.9in, height=0.6in]{ 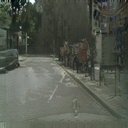}
  \end{minipage} 
      \begin{minipage}[t]{0.13\linewidth} 
    \centering 
    \includegraphics[width=0.9in, height=0.6in]{ 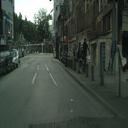}
  \end{minipage} 
\\
  \begin{minipage}[t]{0.13\linewidth} 
    \centering 
    \includegraphics[width=0.9in, height=0.6in]{ 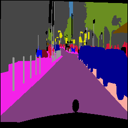}
  \end{minipage} 
      \begin{minipage}[t]{0.13\linewidth} 
    \centering 
    \includegraphics[width=0.9in, height=0.6in]{ 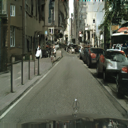}
  \end{minipage} 
    \begin{minipage}[t]{0.13\linewidth} 
    \centering 
    \includegraphics[width=0.9in, height=0.6in]{ 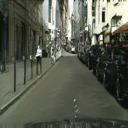}
  \end{minipage} 
    \begin{minipage}[t]{0.13\linewidth} 
    \centering 
    \includegraphics[width=0.9in, height=0.6in]{ 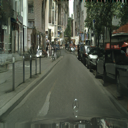}
  \end{minipage} 
    \begin{minipage}[t]{0.13\linewidth} 
    \centering 
    \includegraphics[width=0.9in, height=0.6in]{ 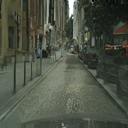}
  \end{minipage} 
    \begin{minipage}[t]{0.13\linewidth} 
    \centering 
    \includegraphics[width=0.9in, height=0.6in]{ 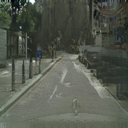}
  \end{minipage} 
      \begin{minipage}[t]{0.13\linewidth} 
    \centering 
    \includegraphics[width=0.9in, height=0.6in]{ 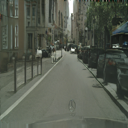}
  \end{minipage} 
\\
  \begin{minipage}[t]{0.13\linewidth} 
    \centering 
    \includegraphics[width=0.9in, height=0.6in]{ 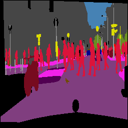}
  \end{minipage} 
      \begin{minipage}[t]{0.13\linewidth} 
    \centering 
    \includegraphics[width=0.9in, height=0.6in]{ 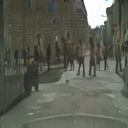}
  \end{minipage} 
    \begin{minipage}[t]{0.13\linewidth} 
    \centering 
    \includegraphics[width=0.9in, height=0.6in]{ 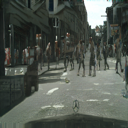}
  \end{minipage} 
    \begin{minipage}[t]{0.13\linewidth} 
    \centering 
    \includegraphics[width=0.9in, height=0.6in]{ 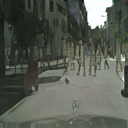}
  \end{minipage} 
    \begin{minipage}[t]{0.13\linewidth} 
    \centering 
    \includegraphics[width=0.9in, height=0.6in]{ 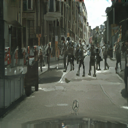}
  \end{minipage} 
    \begin{minipage}[t]{0.13\linewidth} 
    \centering 
    \includegraphics[width=0.9in, height=0.6in]{ 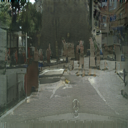}
  \end{minipage} 
      \begin{minipage}[t]{0.13\linewidth} 
    \centering 
    \includegraphics[width=0.9in, height=0.6in]{ 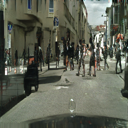}
  \end{minipage} 
\\
  \begin{minipage}[t]{0.13\linewidth} 
    \centering 
    \includegraphics[width=0.9in, height=0.6in]{ 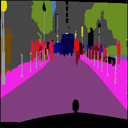}
  \end{minipage} 
      \begin{minipage}[t]{0.13\linewidth} 
    \centering 
    \includegraphics[width=0.9in, height=0.6in]{ 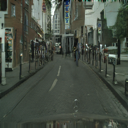}
  \end{minipage} 
    \begin{minipage}[t]{0.13\linewidth} 
    \centering 
    \includegraphics[width=0.9in, height=0.6in]{ 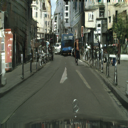}
  \end{minipage} 
    \begin{minipage}[t]{0.13\linewidth} 
    \centering 
    \includegraphics[width=0.9in, height=0.6in]{ 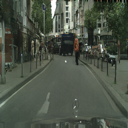}
  \end{minipage} 
    \begin{minipage}[t]{0.13\linewidth} 
    \centering 
    \includegraphics[width=0.9in, height=0.6in]{ 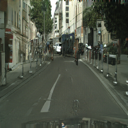}
  \end{minipage} 
    \begin{minipage}[t]{0.13\linewidth} 
    \centering 
    \includegraphics[width=0.9in, height=0.6in]{ 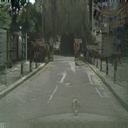}
  \end{minipage} 
      \begin{minipage}[t]{0.13\linewidth} 
    \centering 
    \includegraphics[width=0.9in, height=0.6in]{ 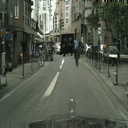}
  \end{minipage} 
\\
  \begin{minipage}[t]{0.13\linewidth} 
    \centering 
    \includegraphics[width=0.9in, height=0.6in]{ 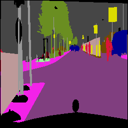}
  \end{minipage} 
      \begin{minipage}[t]{0.13\linewidth} 
    \centering 
    \includegraphics[width=0.9in, height=0.6in]{ 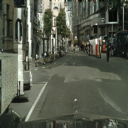}
  \end{minipage} 
    \begin{minipage}[t]{0.13\linewidth} 
    \centering 
    \includegraphics[width=0.9in, height=0.6in]{ 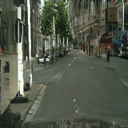}
  \end{minipage} 
    \begin{minipage}[t]{0.13\linewidth} 
    \centering 
    \includegraphics[width=0.9in, height=0.6in]{ 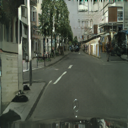}
  \end{minipage} 
    \begin{minipage}[t]{0.13\linewidth} 
    \centering 
    \includegraphics[width=0.9in, height=0.6in]{ 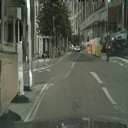}
  \end{minipage} 
    \begin{minipage}[t]{0.13\linewidth} 
    \centering 
    \includegraphics[width=0.9in, height=0.6in]{ 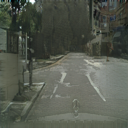}
  \end{minipage} 
      \begin{minipage}[t]{0.13\linewidth} 
    \centering 
    \includegraphics[width=0.9in, height=0.6in]{ 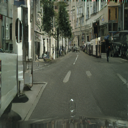}
  \end{minipage} 
  \caption{Qualitative results of ablations on CityScapes task.}
  \label{fig:supp3}
\end{figure*}

%% file: figs/supp4.tex
\begin{figure*}[!tb]
  \begin{minipage}[t]{0.13\linewidth} 
    \centering 
    \text{\small Input}
  \end{minipage} 
    \begin{minipage}[t]{0.13\linewidth} 
    \centering 
        \text{\RomanNumeralCaps{1}}
  \end{minipage} 
    \begin{minipage}[t]{0.13\linewidth} 
    \centering 
        \text{\RomanNumeralCaps{2}}
  \end{minipage} 
    \begin{minipage}[t]{0.13\linewidth} 
    \centering 
        \text{\RomanNumeralCaps{3}}
  \end{minipage} 
    \begin{minipage}[t]{0.13\linewidth} 
    \centering 
            \text{\RomanNumeralCaps{4}}
  \end{minipage} 
    \begin{minipage}[t]{0.13\linewidth} 
    \centering 
            \text{\RomanNumeralCaps{5}}
  \end{minipage} 
      \begin{minipage}[t]{0.13\linewidth} 
    \centering 
    \text{\small DCLGAN}
  \end{minipage} 
  \\
  \begin{minipage}[t]{0.13\linewidth} 
    \centering 
    \includegraphics[width=0.9in, height=0.9in]{ 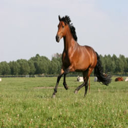}
  \end{minipage} 
      \begin{minipage}[t]{0.13\linewidth} 
    \centering 
    \includegraphics[width=0.9in, height=0.9in]{ 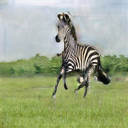}
  \end{minipage} 
    \begin{minipage}[t]{0.13\linewidth} 
    \centering 
    \includegraphics[width=0.9in, height=0.9in]{ 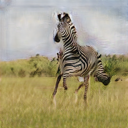}
  \end{minipage} 
    \begin{minipage}[t]{0.13\linewidth} 
    \centering 
    \includegraphics[width=0.9in, height=0.9in]{ 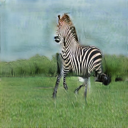}
  \end{minipage} 
    \begin{minipage}[t]{0.13\linewidth} 
    \centering 
    \includegraphics[width=0.9in, height=0.9in]{ 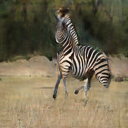}
  \end{minipage} 
    \begin{minipage}[t]{0.13\linewidth} 
    \centering 
    \includegraphics[width=0.9in, height=0.9in]{ 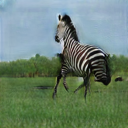}
  \end{minipage} 
      \begin{minipage}[t]{0.13\linewidth} 
    \centering 
    \includegraphics[width=0.9in, height=0.9in]{ 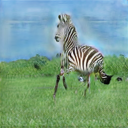}
  \end{minipage} 
\\
  \begin{minipage}[t]{0.13\linewidth} 
    \centering 
    \includegraphics[width=0.9in, height=0.9in]{ 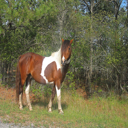}
  \end{minipage} 
      \begin{minipage}[t]{0.13\linewidth} 
    \centering 
    \includegraphics[width=0.9in, height=0.9in]{ 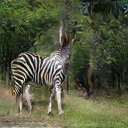}
  \end{minipage} 
    \begin{minipage}[t]{0.13\linewidth} 
    \centering 
    \includegraphics[width=0.9in, height=0.9in]{ 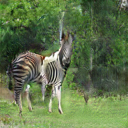}
  \end{minipage} 
    \begin{minipage}[t]{0.13\linewidth} 
    \centering 
    \includegraphics[width=0.9in, height=0.9in]{ 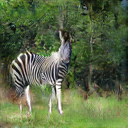}
  \end{minipage} 
    \begin{minipage}[t]{0.13\linewidth} 
    \centering 
    \includegraphics[width=0.9in, height=0.9in]{ 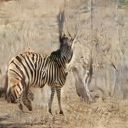}
  \end{minipage} 
    \begin{minipage}[t]{0.13\linewidth} 
    \centering 
    \includegraphics[width=0.9in, height=0.9in]{ 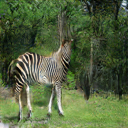}
  \end{minipage} 
      \begin{minipage}[t]{0.13\linewidth} 
    \centering 
    \includegraphics[width=0.9in, height=0.9in]{ 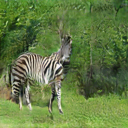}
  \end{minipage} 
\\
  \begin{minipage}[t]{0.13\linewidth} 
    \centering 
    \includegraphics[width=0.9in, height=0.9in]{ 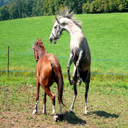}
  \end{minipage} 
      \begin{minipage}[t]{0.13\linewidth} 
    \centering 
    \includegraphics[width=0.9in, height=0.9in]{ 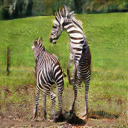}
  \end{minipage} 
    \begin{minipage}[t]{0.13\linewidth} 
    \centering 
    \includegraphics[width=0.9in, height=0.9in]{ 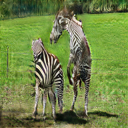}
  \end{minipage} 
    \begin{minipage}[t]{0.13\linewidth} 
    \centering 
    \includegraphics[width=0.9in, height=0.9in]{ 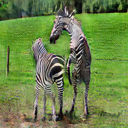}
  \end{minipage} 
    \begin{minipage}[t]{0.13\linewidth} 
    \centering 
    \includegraphics[width=0.9in, height=0.9in]{ 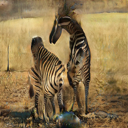}
  \end{minipage} 
    \begin{minipage}[t]{0.13\linewidth} 
    \centering 
    \includegraphics[width=0.9in, height=0.9in]{ 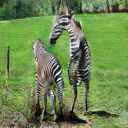}
  \end{minipage} 
      \begin{minipage}[t]{0.13\linewidth} 
    \centering 
    \includegraphics[width=0.9in, height=0.9in]{ 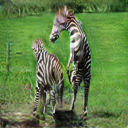}
  \end{minipage} 
\\
  \begin{minipage}[t]{0.13\linewidth} 
    \centering 
    \includegraphics[width=0.9in, height=0.9in]{ 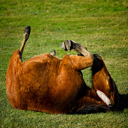}
  \end{minipage} 
      \begin{minipage}[t]{0.13\linewidth} 
    \centering 
    \includegraphics[width=0.9in, height=0.9in]{ 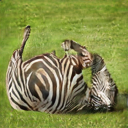}
  \end{minipage} 
    \begin{minipage}[t]{0.13\linewidth} 
    \centering 
    \includegraphics[width=0.9in, height=0.9in]{ 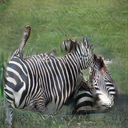}
  \end{minipage} 
    \begin{minipage}[t]{0.13\linewidth} 
    \centering 
    \includegraphics[width=0.9in, height=0.9in]{ 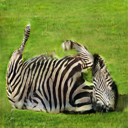}
  \end{minipage} 
    \begin{minipage}[t]{0.13\linewidth} 
    \centering 
    \includegraphics[width=0.9in, height=0.9in]{ 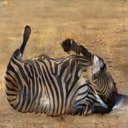}
  \end{minipage} 
    \begin{minipage}[t]{0.13\linewidth} 
    \centering 
    \includegraphics[width=0.9in, height=0.9in]{ 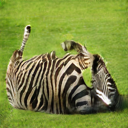}
  \end{minipage} 
      \begin{minipage}[t]{0.13\linewidth} 
    \centering 
    \includegraphics[width=0.9in, height=0.9in]{ 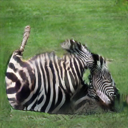}
  \end{minipage} 
\\
  \begin{minipage}[t]{0.13\linewidth} 
    \centering 
    \includegraphics[width=0.9in, height=0.9in]{ 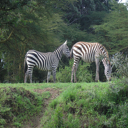}
  \end{minipage} 
      \begin{minipage}[t]{0.13\linewidth} 
    \centering 
    \includegraphics[width=0.9in, height=0.9in]{ 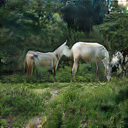}
  \end{minipage} 
    \begin{minipage}[t]{0.13\linewidth} 
    \centering 
    \includegraphics[width=0.9in, height=0.9in]{ 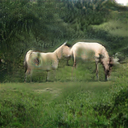}
  \end{minipage} 
    \begin{minipage}[t]{0.13\linewidth} 
    \centering 
    \includegraphics[width=0.9in, height=0.9in]{ 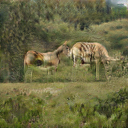}
  \end{minipage} 
    \begin{minipage}[t]{0.13\linewidth} 
    \centering 
    \includegraphics[width=0.9in, height=0.9in]{ 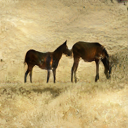}
  \end{minipage} 
    \begin{minipage}[t]{0.13\linewidth} 
    \centering 
    \includegraphics[width=0.9in, height=0.9in]{ 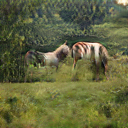}
  \end{minipage} 
      \begin{minipage}[t]{0.13\linewidth} 
    \centering 
    \includegraphics[width=0.9in, height=0.9in]{ 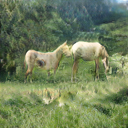}
  \end{minipage} 
\\
  \begin{minipage}[t]{0.13\linewidth} 
    \centering 
    \includegraphics[width=0.9in, height=0.9in]{ 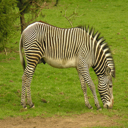}
  \end{minipage} 
      \begin{minipage}[t]{0.13\linewidth} 
    \centering 
    \includegraphics[width=0.9in, height=0.9in]{ 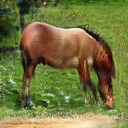}
  \end{minipage} 
    \begin{minipage}[t]{0.13\linewidth} 
    \centering 
    \includegraphics[width=0.9in, height=0.9in]{ 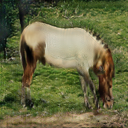}
  \end{minipage} 
    \begin{minipage}[t]{0.13\linewidth} 
    \centering 
    \includegraphics[width=0.9in, height=0.9in]{ 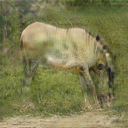}
  \end{minipage} 
    \begin{minipage}[t]{0.13\linewidth} 
    \centering 
    \includegraphics[width=0.9in, height=0.9in]{ 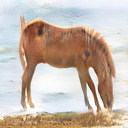}
  \end{minipage} 
    \begin{minipage}[t]{0.13\linewidth} 
    \centering 
    \includegraphics[width=0.9in, height=0.9in]{ 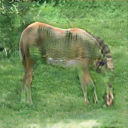}
  \end{minipage} 
      \begin{minipage}[t]{0.13\linewidth} 
    \centering 
    \includegraphics[width=0.9in, height=0.9in]{ 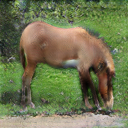}
  \end{minipage} 
  \\
    \begin{minipage}[t]{0.13\linewidth} 
    \centering 
    \includegraphics[width=0.9in, height=0.9in]{ 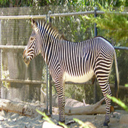}
  \end{minipage} 
      \begin{minipage}[t]{0.13\linewidth} 
    \centering 
    \includegraphics[width=0.9in, height=0.9in]{ 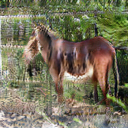}
  \end{minipage} 
    \begin{minipage}[t]{0.13\linewidth} 
    \centering 
    \includegraphics[width=0.9in, height=0.9in]{ 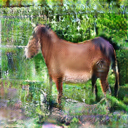}
  \end{minipage} 
    \begin{minipage}[t]{0.13\linewidth} 
    \centering 
    \includegraphics[width=0.9in, height=0.9in]{ 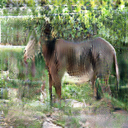}
  \end{minipage} 
    \begin{minipage}[t]{0.13\linewidth} 
    \centering 
    \includegraphics[width=0.9in, height=0.9in]{ 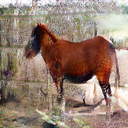}
  \end{minipage} 
    \begin{minipage}[t]{0.13\linewidth} 
    \centering 
    \includegraphics[width=0.9in, height=0.9in]{ 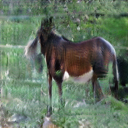}
  \end{minipage} 
      \begin{minipage}[t]{0.13\linewidth} 
    \centering 
    \includegraphics[width=0.9in, height=0.9in]{ 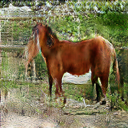}
  \end{minipage} 
  \\
    \begin{minipage}[t]{0.13\linewidth} 
    \centering 
    \includegraphics[width=0.9in, height=0.9in]{ 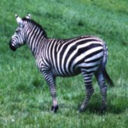}
  \end{minipage} 
      \begin{minipage}[t]{0.13\linewidth} 
    \centering 
    \includegraphics[width=0.9in, height=0.9in]{ 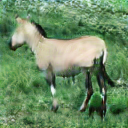}
  \end{minipage} 
    \begin{minipage}[t]{0.13\linewidth} 
    \centering 
    \includegraphics[width=0.9in, height=0.9in]{ 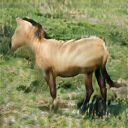}
  \end{minipage} 
    \begin{minipage}[t]{0.13\linewidth} 
    \centering 
    \includegraphics[width=0.9in, height=0.9in]{ 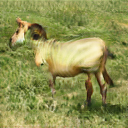}
  \end{minipage} 
    \begin{minipage}[t]{0.13\linewidth} 
    \centering 
    \includegraphics[width=0.9in, height=0.9in]{ 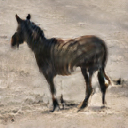}
  \end{minipage} 
    \begin{minipage}[t]{0.13\linewidth} 
    \centering 
    \includegraphics[width=0.9in, height=0.9in]{ 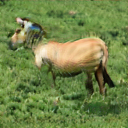}
  \end{minipage} 
      \begin{minipage}[t]{0.13\linewidth} 
    \centering 
    \includegraphics[width=0.9in, height=0.9in]{ 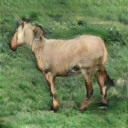}
  \end{minipage} 
  \caption{Qualitative results of ablations on Horse $\rightarrow$ Zebra and Zebra $\rightarrow$ Horse tasks.}
  \label{fig:supp4}
\end{figure*}

%% file: figs/supp5.tex
\begin{figure*}[htb]
  \begin{minipage}[t]{0.095\linewidth} 
    \centering 
    \text{\small Input}
  \end{minipage} 
    \begin{minipage}[t]{0.095\linewidth} 
    \centering 
    \text{\small DCLGAN}
  \end{minipage} 
    \begin{minipage}[t]{0.095\linewidth} 
    \centering 
      \text{\small SimDCL}
  \end{minipage} 
    \begin{minipage}[t]{0.095\linewidth} 
    \centering 
        \text{\small CUT}
  \end{minipage} 
    \begin{minipage}[t]{0.095\linewidth} 
    \centering 
        \text{\small CycleGAN}
  \end{minipage} 
    \begin{minipage}[t]{0.095\linewidth} 
    \centering 
    \text{\small Input}
  \end{minipage} 
    \begin{minipage}[t]{0.095\linewidth} 
    \centering 
    \text{\small DCLGAN}
  \end{minipage} 
    \begin{minipage}[t]{0.095\linewidth} 
    \centering 
      \text{\small SimDCL}
  \end{minipage} 
    \begin{minipage}[t]{0.095\linewidth} 
    \centering 
        \text{\small CUT}
  \end{minipage} 
    \begin{minipage}[t]{0.095\linewidth} 
    \centering 
        \text{\small CycleGAN}
  \end{minipage} 
  \\
  \begin{minipage}[t]{0.095\linewidth} 
    \centering 
    \includegraphics[width=0.62in, height=0.62in]{ 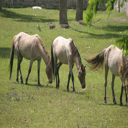}
  \end{minipage} 
    \begin{minipage}[t]{0.095\linewidth} 
    \centering 
    \includegraphics[width=0.62in, height=0.62in]{ 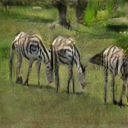}
  \end{minipage} 
      \begin{minipage}[t]{0.095\linewidth} 
    \centering 
    \includegraphics[width=0.62in, height=0.62in]{ 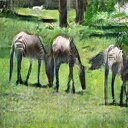}
  \end{minipage} 
    \begin{minipage}[t]{0.095\linewidth} 
    \centering 
    \includegraphics[width=0.62in, height=0.62in]{ 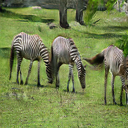}
  \end{minipage} 
    \begin{minipage}[t]{0.095\linewidth} 
    \centering 
    \includegraphics[width=0.62in, height=0.62in]{ 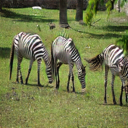}
  \end{minipage} 
  \begin{minipage}[t]{0.095\linewidth} 
    \centering 
    \includegraphics[width=0.62in, height=0.62in]{ 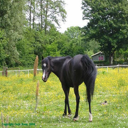}
  \end{minipage} 
    \begin{minipage}[t]{0.095\linewidth} 
    \centering 
    \includegraphics[width=0.62in, height=0.62in]{ 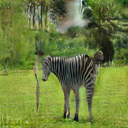}
  \end{minipage} 
      \begin{minipage}[t]{0.095\linewidth} 
    \centering 
    \includegraphics[width=0.62in, height=0.62in]{ 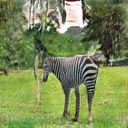}
  \end{minipage} 
    \begin{minipage}[t]{0.095\linewidth} 
    \centering 
    \includegraphics[width=0.62in, height=0.62in]{ 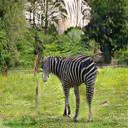}
  \end{minipage} 
    \begin{minipage}[t]{0.095\linewidth} 
    \centering 
    \includegraphics[width=0.62in, height=0.62in]{ 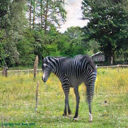}
  \end{minipage} 
\\
  \begin{minipage}[t]{0.095\linewidth} 
    \centering 
    \includegraphics[width=0.62in, height=0.62in]{ 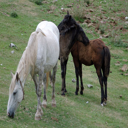}
  \end{minipage} 
    \begin{minipage}[t]{0.095\linewidth} 
    \centering 
    \includegraphics[width=0.62in, height=0.62in]{ 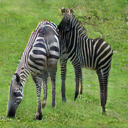}
  \end{minipage} 
      \begin{minipage}[t]{0.095\linewidth} 
    \centering 
    \includegraphics[width=0.62in, height=0.62in]{ 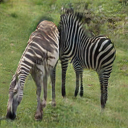}
  \end{minipage} 
    \begin{minipage}[t]{0.095\linewidth} 
    \centering 
    \includegraphics[width=0.62in, height=0.62in]{ 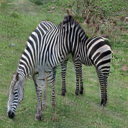}
  \end{minipage} 
    \begin{minipage}[t]{0.095\linewidth} 
    \centering 
    \includegraphics[width=0.62in, height=0.62in]{ 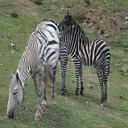}
  \end{minipage} 
  \begin{minipage}[t]{0.095\linewidth} 
    \centering 
    \includegraphics[width=0.62in, height=0.62in]{ 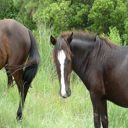}
  \end{minipage} 
    \begin{minipage}[t]{0.095\linewidth} 
    \centering 
    \includegraphics[width=0.62in, height=0.62in]{ 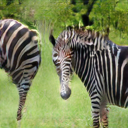}
  \end{minipage} 
      \begin{minipage}[t]{0.095\linewidth} 
    \centering 
    \includegraphics[width=0.62in, height=0.62in]{ 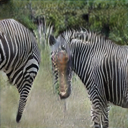}
  \end{minipage} 
    \begin{minipage}[t]{0.095\linewidth} 
    \centering 
    \includegraphics[width=0.62in, height=0.62in]{ 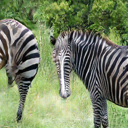}
  \end{minipage} 
    \begin{minipage}[t]{0.095\linewidth} 
    \centering 
    \includegraphics[width=0.62in, height=0.62in]{ 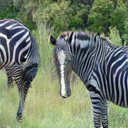}
  \end{minipage} 
\\
  \begin{minipage}[t]{0.095\linewidth} 
    \centering 
    \includegraphics[width=0.62in, height=0.62in]{ 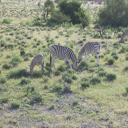}
  \end{minipage} 
    \begin{minipage}[t]{0.095\linewidth} 
    \centering 
    \includegraphics[width=0.62in, height=0.62in]{ 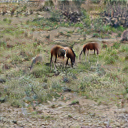}
  \end{minipage} 
      \begin{minipage}[t]{0.095\linewidth} 
    \centering 
    \includegraphics[width=0.62in, height=0.62in]{ 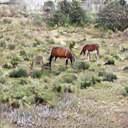}
  \end{minipage} 
    \begin{minipage}[t]{0.095\linewidth} 
    \centering 
    \includegraphics[width=0.62in, height=0.62in]{ 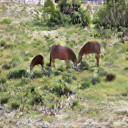}
  \end{minipage} 
    \begin{minipage}[t]{0.095\linewidth} 
    \centering 
    \includegraphics[width=0.62in, height=0.62in]{ 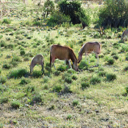}
  \end{minipage} 
  \begin{minipage}[t]{0.095\linewidth} 
    \centering 
    \includegraphics[width=0.62in, height=0.62in]{ 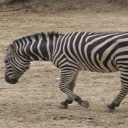}
  \end{minipage} 
    \begin{minipage}[t]{0.095\linewidth} 
    \centering 
    \includegraphics[width=0.62in, height=0.62in]{ 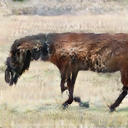}
  \end{minipage} 
      \begin{minipage}[t]{0.095\linewidth} 
    \centering 
    \includegraphics[width=0.62in, height=0.62in]{ 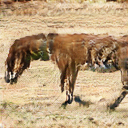}
  \end{minipage} 
    \begin{minipage}[t]{0.095\linewidth} 
    \centering 
    \includegraphics[width=0.62in, height=0.62in]{ 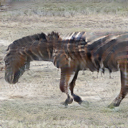}
  \end{minipage} 
    \begin{minipage}[t]{0.095\linewidth} 
    \centering 
    \includegraphics[width=0.62in, height=0.62in]{ 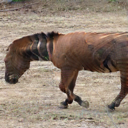}
  \end{minipage} 
\\
  \begin{minipage}[t]{0.095\linewidth} 
    \centering 
    \includegraphics[width=0.62in, height=0.62in]{ 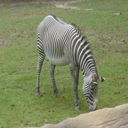}
  \end{minipage} 
    \begin{minipage}[t]{0.095\linewidth} 
    \centering 
    \includegraphics[width=0.62in, height=0.62in]{ 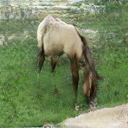}
  \end{minipage} 
      \begin{minipage}[t]{0.095\linewidth} 
    \centering 
    \includegraphics[width=0.62in, height=0.62in]{ 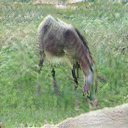}
  \end{minipage} 
    \begin{minipage}[t]{0.095\linewidth} 
    \centering 
    \includegraphics[width=0.62in, height=0.62in]{ 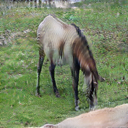}
  \end{minipage} 
    \begin{minipage}[t]{0.095\linewidth} 
    \centering 
    \includegraphics[width=0.62in, height=0.62in]{ 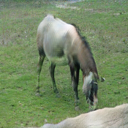}
  \end{minipage} 
  \begin{minipage}[t]{0.095\linewidth} 
    \centering 
    \includegraphics[width=0.62in, height=0.62in]{ 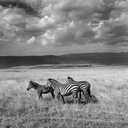}
  \end{minipage} 
    \begin{minipage}[t]{0.095\linewidth} 
    \centering 
    \includegraphics[width=0.62in, height=0.62in]{ 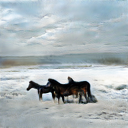}
  \end{minipage} 
      \begin{minipage}[t]{0.095\linewidth} 
    \centering 
    \includegraphics[width=0.62in, height=0.62in]{ 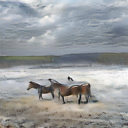}
  \end{minipage} 
    \begin{minipage}[t]{0.095\linewidth} 
    \centering 
    \includegraphics[width=0.62in, height=0.62in]{ 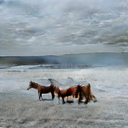}
  \end{minipage} 
    \begin{minipage}[t]{0.095\linewidth} 
    \centering 
    \includegraphics[width=0.62in, height=0.62in]{ 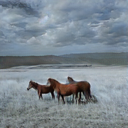}
  \end{minipage} 
  \caption{Additional results of Horse $\leftrightarrow$ Zebra.}
  \label{fig:supp5}
\end{figure*}

%% file: figs/supp6.tex
\begin{figure*}[htb]
  \begin{minipage}[t]{0.095\linewidth} 
    \centering 
    \text{\small Input}
  \end{minipage} 
    \begin{minipage}[t]{0.095\linewidth} 
    \centering 
    \text{\small DCLGAN}
  \end{minipage} 
    \begin{minipage}[t]{0.095\linewidth} 
    \centering 
      \text{\small SimDCL}
  \end{minipage} 
    \begin{minipage}[t]{0.095\linewidth} 
    \centering 
        \text{\small CUT}
  \end{minipage} 
    \begin{minipage}[t]{0.095\linewidth} 
    \centering 
        \text{\small CycleGAN}
  \end{minipage} 
    \begin{minipage}[t]{0.095\linewidth} 
    \centering 
    \text{\small Input}
  \end{minipage} 
    \begin{minipage}[t]{0.095\linewidth} 
    \centering 
    \text{\small DCLGAN}
  \end{minipage} 
    \begin{minipage}[t]{0.095\linewidth} 
    \centering 
      \text{\small SimDCL}
  \end{minipage} 
    \begin{minipage}[t]{0.095\linewidth} 
    \centering 
        \text{\small CUT}
  \end{minipage} 
    \begin{minipage}[t]{0.095\linewidth} 
    \centering 
        \text{\small CycleGAN}
  \end{minipage} 
  \\
  \begin{minipage}[t]{0.095\linewidth} 
    \centering 
    \includegraphics[width=0.62in, height=0.62in]{ 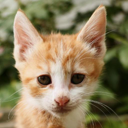}
  \end{minipage} 
    \begin{minipage}[t]{0.095\linewidth} 
    \centering 
    \includegraphics[width=0.62in, height=0.62in]{ 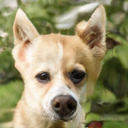}
  \end{minipage} 
      \begin{minipage}[t]{0.095\linewidth} 
    \centering 
    \includegraphics[width=0.62in, height=0.62in]{ 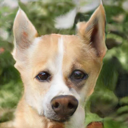}
  \end{minipage} 
    \begin{minipage}[t]{0.095\linewidth} 
    \centering 
    \includegraphics[width=0.62in, height=0.62in]{ 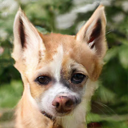}
  \end{minipage} 
    \begin{minipage}[t]{0.095\linewidth} 
    \centering 
    \includegraphics[width=0.62in, height=0.62in]{ 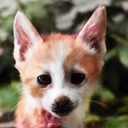}
  \end{minipage} 
  \begin{minipage}[t]{0.095\linewidth} 
    \centering 
    \includegraphics[width=0.62in, height=0.62in]{ 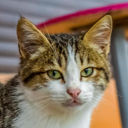}
  \end{minipage} 
    \begin{minipage}[t]{0.095\linewidth} 
    \centering 
    \includegraphics[width=0.62in, height=0.62in]{ 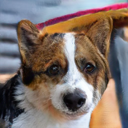}
  \end{minipage} 
      \begin{minipage}[t]{0.095\linewidth} 
    \centering 
    \includegraphics[width=0.62in, height=0.62in]{ 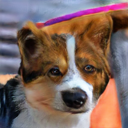}
  \end{minipage} 
    \begin{minipage}[t]{0.095\linewidth} 
    \centering 
    \includegraphics[width=0.62in, height=0.62in]{ 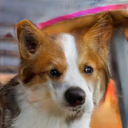}
  \end{minipage} 
    \begin{minipage}[t]{0.095\linewidth} 
    \centering 
    \includegraphics[width=0.62in, height=0.62in]{ 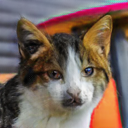}
  \end{minipage} 
\\
  \begin{minipage}[t]{0.095\linewidth} 
    \centering 
    \includegraphics[width=0.62in, height=0.62in]{ 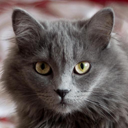}
  \end{minipage} 
    \begin{minipage}[t]{0.095\linewidth} 
    \centering 
    \includegraphics[width=0.62in, height=0.62in]{ 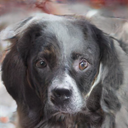}
  \end{minipage} 
      \begin{minipage}[t]{0.095\linewidth} 
    \centering 
    \includegraphics[width=0.62in, height=0.62in]{ 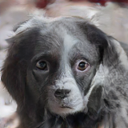}
  \end{minipage} 
    \begin{minipage}[t]{0.095\linewidth} 
    \centering 
    \includegraphics[width=0.62in, height=0.62in]{ 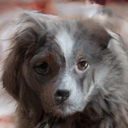}
  \end{minipage} 
    \begin{minipage}[t]{0.095\linewidth} 
    \centering 
    \includegraphics[width=0.62in, height=0.62in]{ 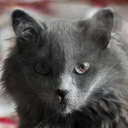}
  \end{minipage} 
  \begin{minipage}[t]{0.095\linewidth} 
    \centering 
    \includegraphics[width=0.62in, height=0.62in]{ 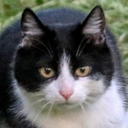}
  \end{minipage} 
    \begin{minipage}[t]{0.095\linewidth} 
    \centering 
    \includegraphics[width=0.62in, height=0.62in]{ 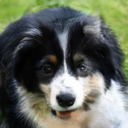}
  \end{minipage} 
      \begin{minipage}[t]{0.095\linewidth} 
    \centering 
    \includegraphics[width=0.62in, height=0.62in]{ 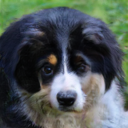}
  \end{minipage} 
    \begin{minipage}[t]{0.095\linewidth} 
    \centering 
    \includegraphics[width=0.62in, height=0.62in]{ 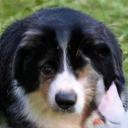}
  \end{minipage} 
    \begin{minipage}[t]{0.095\linewidth} 
    \centering 
    \includegraphics[width=0.62in, height=0.62in]{ 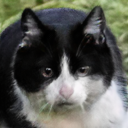}
  \end{minipage} 
\\
  \begin{minipage}[t]{0.095\linewidth} 
    \centering 
    \includegraphics[width=0.62in, height=0.62in]{ 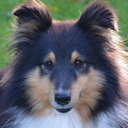}
  \end{minipage} 
    \begin{minipage}[t]{0.095\linewidth} 
    \centering 
    \includegraphics[width=0.62in, height=0.62in]{ 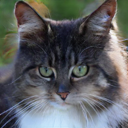}
  \end{minipage} 
      \begin{minipage}[t]{0.095\linewidth} 
    \centering 
    \includegraphics[width=0.62in, height=0.62in]{ 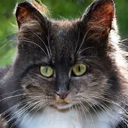}
  \end{minipage} 
    \begin{minipage}[t]{0.095\linewidth} 
    \centering 
    \includegraphics[width=0.62in, height=0.62in]{ 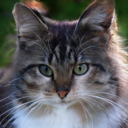}
  \end{minipage} 
    \begin{minipage}[t]{0.095\linewidth} 
    \centering 
    \includegraphics[width=0.62in, height=0.62in]{ 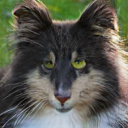}
  \end{minipage} 
  \begin{minipage}[t]{0.095\linewidth} 
    \centering 
    \includegraphics[width=0.62in, height=0.62in]{ 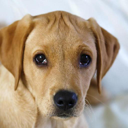}
  \end{minipage} 
    \begin{minipage}[t]{0.095\linewidth} 
    \centering 
    \includegraphics[width=0.62in, height=0.62in]{ 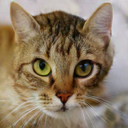}
  \end{minipage} 
      \begin{minipage}[t]{0.095\linewidth} 
    \centering 
    \includegraphics[width=0.62in, height=0.62in]{ 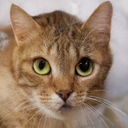}
  \end{minipage} 
    \begin{minipage}[t]{0.095\linewidth} 
    \centering 
    \includegraphics[width=0.62in, height=0.62in]{ 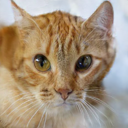}
  \end{minipage} 
    \begin{minipage}[t]{0.095\linewidth} 
    \centering 
    \includegraphics[width=0.62in, height=0.62in]{ 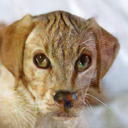}
  \end{minipage} 
\\
  \begin{minipage}[t]{0.095\linewidth} 
    \centering 
    \includegraphics[width=0.62in, height=0.62in]{ 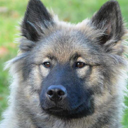}
  \end{minipage} 
    \begin{minipage}[t]{0.095\linewidth} 
    \centering 
    \includegraphics[width=0.62in, height=0.62in]{ 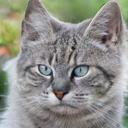}
  \end{minipage} 
      \begin{minipage}[t]{0.095\linewidth} 
    \centering 
    \includegraphics[width=0.62in, height=0.62in]{ 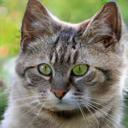}
  \end{minipage} 
    \begin{minipage}[t]{0.095\linewidth} 
    \centering 
    \includegraphics[width=0.62in, height=0.62in]{ 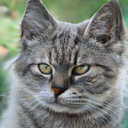}
  \end{minipage} 
    \begin{minipage}[t]{0.095\linewidth} 
    \centering 
    \includegraphics[width=0.62in, height=0.62in]{ 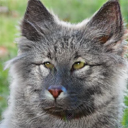}
  \end{minipage} 
  \begin{minipage}[t]{0.095\linewidth} 
    \centering 
    \includegraphics[width=0.62in, height=0.62in]{ 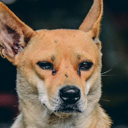}
  \end{minipage} 
    \begin{minipage}[t]{0.095\linewidth} 
    \centering 
    \includegraphics[width=0.62in, height=0.62in]{ 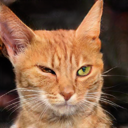}
  \end{minipage} 
      \begin{minipage}[t]{0.095\linewidth} 
    \centering 
    \includegraphics[width=0.62in, height=0.62in]{ 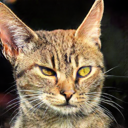}
  \end{minipage} 
    \begin{minipage}[t]{0.095\linewidth} 
    \centering 
    \includegraphics[width=0.62in, height=0.62in]{ 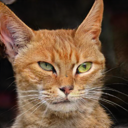}
  \end{minipage} 
    \begin{minipage}[t]{0.095\linewidth} 
    \centering 
    \includegraphics[width=0.62in, height=0.62in]{ 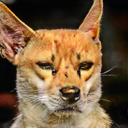}
  \end{minipage} 
  \caption{Additional results of Cat $\leftrightarrow$ Dog.}
  \label{fig:supp6}
\end{figure*}

%% file: figs/supp7.tex
\begin{figure*}[htb]
  \begin{minipage}[t]{0.095\linewidth} 
    \centering 
    \text{\small Input}
  \end{minipage} 
    \begin{minipage}[t]{0.095\linewidth} 
    \centering 
    \text{\small DCLGAN}
  \end{minipage} 
    \begin{minipage}[t]{0.095\linewidth} 
    \centering 
      \text{\small SimDCL}
  \end{minipage} 
    \begin{minipage}[t]{0.095\linewidth} 
    \centering 
        \text{\small CUT}
  \end{minipage} 
    \begin{minipage}[t]{0.095\linewidth} 
    \centering 
        \text{\small CycleGAN}
  \end{minipage} 
    \begin{minipage}[t]{0.095\linewidth} 
    \centering 
    \text{\small Input}
  \end{minipage} 
    \begin{minipage}[t]{0.095\linewidth} 
    \centering 
    \text{\small DCLGAN}
  \end{minipage} 
    \begin{minipage}[t]{0.095\linewidth} 
    \centering 
      \text{\small SimDCL}
  \end{minipage} 
    \begin{minipage}[t]{0.095\linewidth} 
    \centering 
        \text{\small CUT}
  \end{minipage} 
    \begin{minipage}[t]{0.095\linewidth} 
    \centering 
        \text{\small CycleGAN}
  \end{minipage} 
  \\
  \begin{minipage}[t]{0.095\linewidth} 
    \centering 
    \includegraphics[width=0.62in, height=0.62in]{ 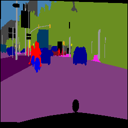}
  \end{minipage} 
    \begin{minipage}[t]{0.095\linewidth} 
    \centering 
    \includegraphics[width=0.62in, height=0.62in]{ 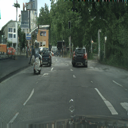}
  \end{minipage} 
      \begin{minipage}[t]{0.095\linewidth} 
    \centering 
    \includegraphics[width=0.62in, height=0.62in]{ 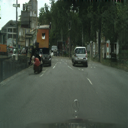}
  \end{minipage} 
    \begin{minipage}[t]{0.095\linewidth} 
    \centering 
    \includegraphics[width=0.62in, height=0.62in]{ 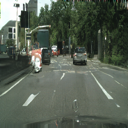}
  \end{minipage} 
    \begin{minipage}[t]{0.095\linewidth} 
    \centering 
    \includegraphics[width=0.62in, height=0.62in]{ 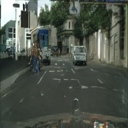}
  \end{minipage} 
  \begin{minipage}[t]{0.095\linewidth} 
    \centering 
    \includegraphics[width=0.62in, height=0.62in]{ 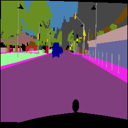}
  \end{minipage} 
    \begin{minipage}[t]{0.095\linewidth} 
    \centering 
    \includegraphics[width=0.62in, height=0.62in]{ 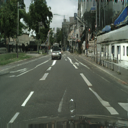}
  \end{minipage} 
      \begin{minipage}[t]{0.095\linewidth} 
    \centering 
    \includegraphics[width=0.62in, height=0.62in]{ 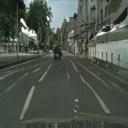}
  \end{minipage} 
    \begin{minipage}[t]{0.095\linewidth} 
    \centering 
    \includegraphics[width=0.62in, height=0.62in]{ 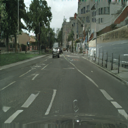}
  \end{minipage} 
    \begin{minipage}[t]{0.095\linewidth} 
    \centering 
    \includegraphics[width=0.62in, height=0.62in]{ 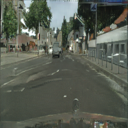}
  \end{minipage} 
\\
  \begin{minipage}[t]{0.095\linewidth} 
    \centering 
    \includegraphics[width=0.62in, height=0.62in]{ 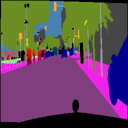}
  \end{minipage} 
    \begin{minipage}[t]{0.095\linewidth} 
    \centering 
    \includegraphics[width=0.62in, height=0.62in]{ 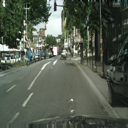}
  \end{minipage} 
      \begin{minipage}[t]{0.095\linewidth} 
    \centering 
    \includegraphics[width=0.62in, height=0.62in]{ 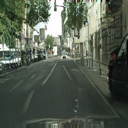}
  \end{minipage} 
    \begin{minipage}[t]{0.095\linewidth} 
    \centering 
    \includegraphics[width=0.62in, height=0.62in]{ 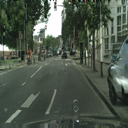}
  \end{minipage} 
    \begin{minipage}[t]{0.095\linewidth} 
    \centering 
    \includegraphics[width=0.62in, height=0.62in]{ 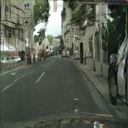}
  \end{minipage} 
  \begin{minipage}[t]{0.095\linewidth} 
    \centering 
    \includegraphics[width=0.62in, height=0.62in]{ 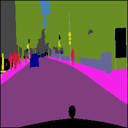}
  \end{minipage} 
    \begin{minipage}[t]{0.095\linewidth} 
    \centering 
    \includegraphics[width=0.62in, height=0.62in]{ 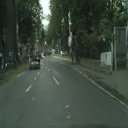}
  \end{minipage} 
      \begin{minipage}[t]{0.095\linewidth} 
    \centering 
    \includegraphics[width=0.62in, height=0.62in]{ 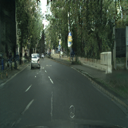}
  \end{minipage} 
    \begin{minipage}[t]{0.095\linewidth} 
    \centering 
    \includegraphics[width=0.62in, height=0.62in]{ 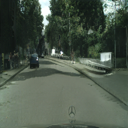}
  \end{minipage} 
    \begin{minipage}[t]{0.095\linewidth} 
    \centering 
    \includegraphics[width=0.62in, height=0.62in]{ 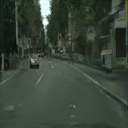}
  \end{minipage} 
\\
  \begin{minipage}[t]{0.095\linewidth} 
    \centering 
    \includegraphics[width=0.62in, height=0.62in]{ 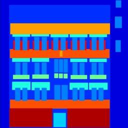}
  \end{minipage} 
    \begin{minipage}[t]{0.095\linewidth} 
    \centering 
    \includegraphics[width=0.62in, height=0.62in]{ 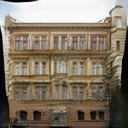}
  \end{minipage} 
      \begin{minipage}[t]{0.095\linewidth} 
    \centering 
    \includegraphics[width=0.62in, height=0.62in]{ 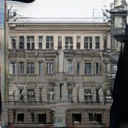}
  \end{minipage} 
    \begin{minipage}[t]{0.095\linewidth} 
    \centering 
    \includegraphics[width=0.62in, height=0.62in]{ 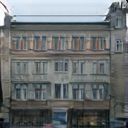}
  \end{minipage} 
    \begin{minipage}[t]{0.095\linewidth} 
    \centering 
    \includegraphics[width=0.62in, height=0.62in]{ 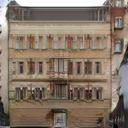}
  \end{minipage} 
  \begin{minipage}[t]{0.095\linewidth} 
    \centering 
    \includegraphics[width=0.62in, height=0.62in]{ 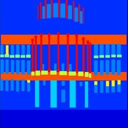}
  \end{minipage} 
    \begin{minipage}[t]{0.095\linewidth} 
    \centering 
    \includegraphics[width=0.62in, height=0.62in]{ 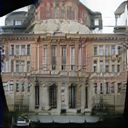}
  \end{minipage} 
      \begin{minipage}[t]{0.095\linewidth} 
    \centering 
    \includegraphics[width=0.62in, height=0.62in]{ 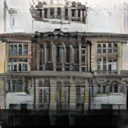}
  \end{minipage} 
    \begin{minipage}[t]{0.095\linewidth} 
    \centering 
    \includegraphics[width=0.62in, height=0.62in]{ 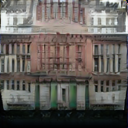}
  \end{minipage} 
    \begin{minipage}[t]{0.095\linewidth} 
    \centering 
    \includegraphics[width=0.62in, height=0.62in]{ 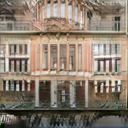}
  \end{minipage} 
\\
  \begin{minipage}[t]{0.095\linewidth} 
    \centering 
    \includegraphics[width=0.62in, height=0.62in]{ 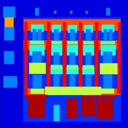}
  \end{minipage} 
    \begin{minipage}[t]{0.095\linewidth} 
    \centering 
    \includegraphics[width=0.62in, height=0.62in]{ 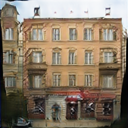}
  \end{minipage} 
      \begin{minipage}[t]{0.095\linewidth} 
    \centering 
    \includegraphics[width=0.62in, height=0.62in]{ 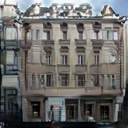}
  \end{minipage} 
    \begin{minipage}[t]{0.095\linewidth} 
    \centering 
    \includegraphics[width=0.62in, height=0.62in]{ 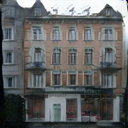}
  \end{minipage} 
    \begin{minipage}[t]{0.095\linewidth} 
    \centering 
    \includegraphics[width=0.62in, height=0.62in]{ 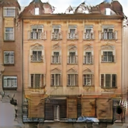}
  \end{minipage} 
  \begin{minipage}[t]{0.095\linewidth} 
    \centering 
    \includegraphics[width=0.62in, height=0.62in]{ 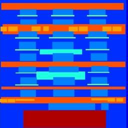}
  \end{minipage} 
    \begin{minipage}[t]{0.095\linewidth} 
    \centering 
    \includegraphics[width=0.62in, height=0.62in]{ 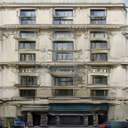}
  \end{minipage} 
      \begin{minipage}[t]{0.095\linewidth} 
    \centering 
    \includegraphics[width=0.62in, height=0.62in]{ 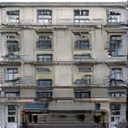}
  \end{minipage} 
    \begin{minipage}[t]{0.095\linewidth} 
    \centering 
    \includegraphics[width=0.62in, height=0.62in]{ 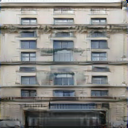}
  \end{minipage} 
    \begin{minipage}[t]{0.095\linewidth} 
    \centering 
    \includegraphics[width=0.62in, height=0.62in]{ 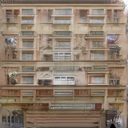}
  \end{minipage} 
  \caption{Additional results of CityScapes and Label $\rightarrow$ Facade.}
  \label{fig:supp7}
\end{figure*}

%% file: figs/supp8.tex
\begin{figure*}[htb]
  \begin{minipage}[t]{0.095\linewidth} 
    \centering 
    \text{\small Input}
  \end{minipage} 
    \begin{minipage}[t]{0.095\linewidth} 
    \centering 
    \text{\small DCLGAN}
  \end{minipage} 
    \begin{minipage}[t]{0.095\linewidth} 
    \centering 
      \text{\small SimDCL}
  \end{minipage} 
    \begin{minipage}[t]{0.095\linewidth} 
    \centering 
        \text{\small CUT}
  \end{minipage} 
    \begin{minipage}[t]{0.095\linewidth} 
    \centering 
        \text{\small CycleGAN}
  \end{minipage} 
    \begin{minipage}[t]{0.095\linewidth} 
    \centering 
    \text{\small Input}
  \end{minipage} 
    \begin{minipage}[t]{0.095\linewidth} 
    \centering 
    \text{\small DCLGAN}
  \end{minipage} 
    \begin{minipage}[t]{0.095\linewidth} 
    \centering 
      \text{\small SimDCL}
  \end{minipage} 
    \begin{minipage}[t]{0.095\linewidth} 
    \centering 
        \text{\small CUT}
  \end{minipage} 
    \begin{minipage}[t]{0.095\linewidth} 
    \centering 
        \text{\small CycleGAN}
  \end{minipage} 
  \\
  \begin{minipage}[t]{0.095\linewidth} 
    \centering 
    \includegraphics[width=0.62in, height=0.62in]{ 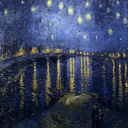}
  \end{minipage} 
    \begin{minipage}[t]{0.095\linewidth} 
    \centering 
    \includegraphics[width=0.62in, height=0.62in]{ 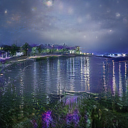}
  \end{minipage} 
      \begin{minipage}[t]{0.095\linewidth} 
    \centering 
    \includegraphics[width=0.62in, height=0.62in]{ 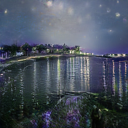}
  \end{minipage} 
    \begin{minipage}[t]{0.095\linewidth} 
    \centering 
    \includegraphics[width=0.62in, height=0.62in]{ 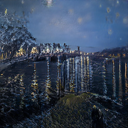}
  \end{minipage} 
    \begin{minipage}[t]{0.095\linewidth} 
    \centering 
    \includegraphics[width=0.62in, height=0.62in]{ 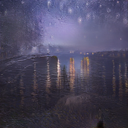}
  \end{minipage} 
  \begin{minipage}[t]{0.095\linewidth} 
    \centering 
    \includegraphics[width=0.62in, height=0.62in]{ 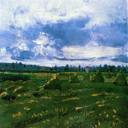}
  \end{minipage} 
    \begin{minipage}[t]{0.095\linewidth} 
    \centering 
    \includegraphics[width=0.62in, height=0.62in]{ 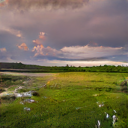}
  \end{minipage} 
      \begin{minipage}[t]{0.095\linewidth} 
    \centering 
    \includegraphics[width=0.62in, height=0.62in]{ 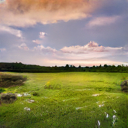}
  \end{minipage} 
    \begin{minipage}[t]{0.095\linewidth} 
    \centering 
    \includegraphics[width=0.62in, height=0.62in]{ 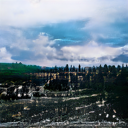}
  \end{minipage} 
    \begin{minipage}[t]{0.095\linewidth} 
    \centering 
    \includegraphics[width=0.62in, height=0.62in]{ 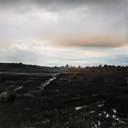}
  \end{minipage} 
\\
  \begin{minipage}[t]{0.095\linewidth} 
    \centering 
    \includegraphics[width=0.62in, height=0.62in]{ 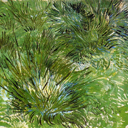}
  \end{minipage} 
    \begin{minipage}[t]{0.095\linewidth} 
    \centering 
    \includegraphics[width=0.62in, height=0.62in]{ 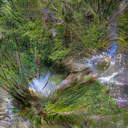}
  \end{minipage} 
      \begin{minipage}[t]{0.095\linewidth} 
    \centering 
    \includegraphics[width=0.62in, height=0.62in]{ 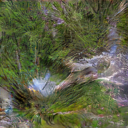}
  \end{minipage} 
    \begin{minipage}[t]{0.095\linewidth} 
    \centering 
    \includegraphics[width=0.62in, height=0.62in]{ 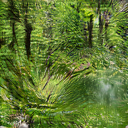}
  \end{minipage} 
    \begin{minipage}[t]{0.095\linewidth} 
    \centering 
    \includegraphics[width=0.62in, height=0.62in]{ 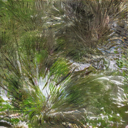}
  \end{minipage} 
  \begin{minipage}[t]{0.095\linewidth} 
    \centering 
    \includegraphics[width=0.62in, height=0.62in]{ 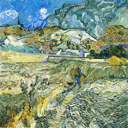}
  \end{minipage} 
    \begin{minipage}[t]{0.095\linewidth} 
    \centering 
    \includegraphics[width=0.62in, height=0.62in]{ 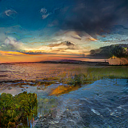}
  \end{minipage} 
      \begin{minipage}[t]{0.095\linewidth} 
    \centering 
    \includegraphics[width=0.62in, height=0.62in]{ 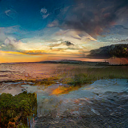}
  \end{minipage} 
    \begin{minipage}[t]{0.095\linewidth} 
    \centering 
    \includegraphics[width=0.62in, height=0.62in]{ 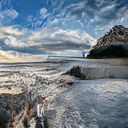}
  \end{minipage} 
    \begin{minipage}[t]{0.095\linewidth} 
    \centering 
    \includegraphics[width=0.62in, height=0.62in]{ 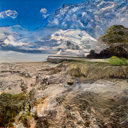}
  \end{minipage} 
\\
  \begin{minipage}[t]{0.095\linewidth} 
    \centering 
    \includegraphics[width=0.62in, height=0.62in]{ 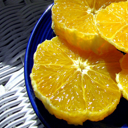}
  \end{minipage} 
    \begin{minipage}[t]{0.095\linewidth} 
    \centering 
    \includegraphics[width=0.62in, height=0.62in]{ 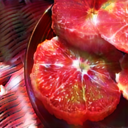}
  \end{minipage} 
      \begin{minipage}[t]{0.095\linewidth} 
    \centering 
    \includegraphics[width=0.62in, height=0.62in]{ 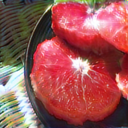}
  \end{minipage} 
    \begin{minipage}[t]{0.095\linewidth} 
    \centering 
    \includegraphics[width=0.62in, height=0.62in]{ 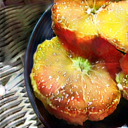}
  \end{minipage} 
    \begin{minipage}[t]{0.095\linewidth} 
    \centering 
    \includegraphics[width=0.62in, height=0.62in]{ 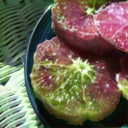}
  \end{minipage} 
  \begin{minipage}[t]{0.095\linewidth} 
    \centering 
    \includegraphics[width=0.62in, height=0.62in]{ 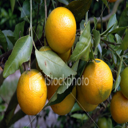}
  \end{minipage} 
    \begin{minipage}[t]{0.095\linewidth} 
    \centering 
    \includegraphics[width=0.62in, height=0.62in]{ 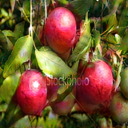}
  \end{minipage} 
      \begin{minipage}[t]{0.095\linewidth} 
    \centering 
    \includegraphics[width=0.62in, height=0.62in]{ 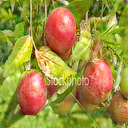}
  \end{minipage} 
    \begin{minipage}[t]{0.095\linewidth} 
    \centering 
    \includegraphics[width=0.62in, height=0.62in]{ 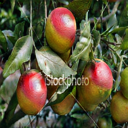}
  \end{minipage} 
    \begin{minipage}[t]{0.095\linewidth} 
    \centering 
    \includegraphics[width=0.62in, height=0.62in]{ 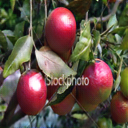}
  \end{minipage} 
\\
  \begin{minipage}[t]{0.095\linewidth} 
    \centering 
    \includegraphics[width=0.62in, height=0.62in]{ 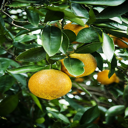}
  \end{minipage} 
    \begin{minipage}[t]{0.095\linewidth} 
    \centering 
    \includegraphics[width=0.62in, height=0.62in]{ 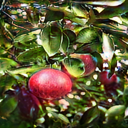}
  \end{minipage} 
      \begin{minipage}[t]{0.095\linewidth} 
    \centering 
    \includegraphics[width=0.62in, height=0.62in]{ 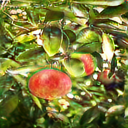}
  \end{minipage} 
    \begin{minipage}[t]{0.095\linewidth} 
    \centering 
    \includegraphics[width=0.62in, height=0.62in]{ 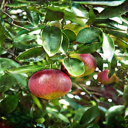}
  \end{minipage} 
    \begin{minipage}[t]{0.095\linewidth} 
    \centering 
    \includegraphics[width=0.62in, height=0.62in]{ 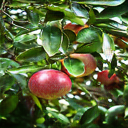}
  \end{minipage} 
  \begin{minipage}[t]{0.095\linewidth} 
    \centering 
    \includegraphics[width=0.62in, height=0.62in]{ 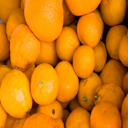}
  \end{minipage} 
    \begin{minipage}[t]{0.095\linewidth} 
    \centering 
    \includegraphics[width=0.62in, height=0.62in]{ 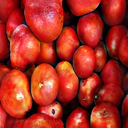}
  \end{minipage} 
      \begin{minipage}[t]{0.095\linewidth} 
    \centering 
    \includegraphics[width=0.62in, height=0.62in]{ 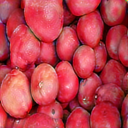}
  \end{minipage} 
    \begin{minipage}[t]{0.095\linewidth} 
    \centering 
    \includegraphics[width=0.62in, height=0.62in]{ 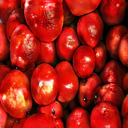}
  \end{minipage} 
    \begin{minipage}[t]{0.095\linewidth} 
    \centering 
    \includegraphics[width=0.62in, height=0.62in]{ 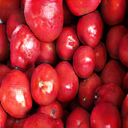}
  \end{minipage} 
  \caption{Additional results of Van Gogh $\rightarrow$ Photo and Orange $\rightarrow$ Apple.}
  \label{fig:supp8}
\end{figure*}